\documentclass[twoside]{article}
\usepackage[round]{natbib}
\usepackage{times}
\usepackage{epsfig}
\usepackage{graphicx}
\usepackage{amsmath}
\usepackage{amssymb}
\usepackage{enumerate}
\usepackage{amsfonts,lscape,color,url}
\usepackage{adjustbox}
\usepackage{graphicx,graphics,wrapfig,epstopdf,subfig}
\usepackage{algorithm}
\usepackage{algorithmic}
\newcommand{\beq}{\begin{equation}}
\newcommand{\eeq}{\end{equation}}
\newcommand{\beqs}{\begin{eqnarray}}
\newcommand{\eeqs}{\end{eqnarray}}
\newcommand{\barr}{\begin{array}}
\newcommand{\earr}{\end{array}}

\newcommand{\eg}{\textit{e.g.}}
\newcommand{\ie}{\textit{i.e.}}

\newcommand{\Amat}{{\bf A}}
\newcommand{\Bmat}{{\bf B}}
\newcommand{\Cmat}{{\bf C}}
\newcommand{\Dmat}{{\bf D}}
\newcommand{\Emat}{{\bf E}}
\newcommand{\Fmat}{{\bf F}}

\newcommand{\Imat}{{\bf I}}

\newcommand{\Smat}{{\bf S}}

\newcommand{\Wmat}{{\bf W}}
\newcommand{\Xmat}{{\bf X}}
\newcommand{\Ymat}{{\bf Y}}
\newcommand{\Zmat}{{\bf Z}}

\newcommand{\ev}{{\boldsymbol e}}

\newcommand{\sv}{{\boldsymbol s}}

\newcommand{\uv}{{\boldsymbol u}}
\newcommand{\vv}{{\boldsymbol v}}

\newcommand{\yv}{{\boldsymbol y}}

\newcommand{\zv}{{\boldsymbol z}}

\newcommand{\Ximat}{{\boldsymbol \Xi}}

\newcommand{\Sigmamat}{{\boldsymbol \Sigma}}

\newcommand{\Phimat}{{\boldsymbol \Phi}}
\newcommand{\Psimat}{{\boldsymbol \Psi}}

\newcommand{\Lambdamat}{{\boldsymbol \Lambda}}

\newcommand{\betav}{{\boldsymbol \beta}}
\newcommand{\gammav}{{\boldsymbol \gamma}}
\newcommand{\deltav}{{\boldsymbol \delta}}

\newcommand{\thetav}{{\boldsymbol \theta}}

\newcommand{\lambdav}{{\boldsymbol \lambda}}

\usepackage[accepted]{aistats2016}
\begin{document}

%

%

\twocolumn[

\aistatstitle{A Deep Generative Deconvolutional Image Model}

\aistatsauthor{ Yunchen Pu \And Xin Yuan \And Andrew Stevens \And Chunyuan Li \And Lawrence Carin }

\aistatsaddress{Duke University \And Bell Labs \And Duke University \And Duke University \And Duke University } ]

\begin{abstract}
A deep generative model is developed for representation and analysis of images, based on a hierarchical convolutional dictionary-learning framework. Stochastic {\em unpooling} is employed to link consecutive layers in the model, yielding top-down image generation.
A Bayesian support vector machine is linked to the top-layer features, yielding max-margin discrimination.  Deep deconvolutional inference is employed when testing, to infer the latent features, and the top-layer features are connected with the max-margin classifier for discrimination tasks. The model is efficiently trained using a Monte Carlo expectation-maximization (MCEM) algorithm, with implementation on graphical processor units (GPUs) for efficient large-scale learning, and fast testing.
Excellent results are obtained on several benchmark datasets, including ImageNet, demonstrating that the proposed model achieves results that are highly competitive with similarly sized convolutional neural networks.
\end{abstract}

\section{Introduction}

Convolutional neural networks (CNN) \citep{LeCun89CN} are effective tools for image and video analysis \citep{return,HintonNIPS2012,Atari,sermanet2013overfeat}. The CNN is characterized by feedforward (bottom-up) sequential application of convolutional filterbanks, pointwise nonlinear functions (e.g., sigmoid or hyperbolic tangent), and pooling.
Supervision in CNN is typically implemented via a fully-connected layer at the top of the deep architecture, usually with a softmax classifier \citep{Ciresan11IJCAI,HeSSPnet,Jarrett09ICCV,HintonNIPS2012}. 


A parallel line of research concerns dictionary learning \citep{Mairal08superviseddictionary,zhang10CVPR,Zhou12TIP} based on a set of image {\em patches}. In this setting one imposes sparsity constraints on the dictionary weights with which the data are represented. For image analysis/processing tasks, rather than using a patch-based model, there has been recent interest in deconvolutional networks (DN) \citep{Chen2011ICML,Chen13deepCFA,Zeiler10CVPR}.
In a DN one uses dictionary learning on an entire image (as opposed to the patches of an image), and each dictionary element is convolved with a sparse set of weights that exist across the entire image.
Such models are termed ``deconvolutional'' because, given a learned dictionary, the features at test are found through deconvolution.
One may build deep deconvolutional models, which typically employ a pooling step like the CNN \citep{Chen2011ICML,Chen13deepCFA}. The convolutional filterbank of the CNN is replaced in the DN by a library of convolutional dictionaries.


In this paper we develop a new deep generative model for images, based on convolutional dictionary learning. At test, after the dictionary elements are learned, deconvolutional inference is employed, like in the aforementioned DN research. The proposed method is related to \citet{Chen2011ICML,Chen13deepCFA}, but a complete top-down generative model is developed, with stochastic {\em unpooling} connecting model layers (this is distinct from almost all other models, which employ bottom-up pooling). \citet{Chen2011ICML,Chen13deepCFA} trained each layer separately, sequentially, with no final coupling of the overall model (significantly undermining classification performance). Further, in \citet{Chen2011ICML,Chen13deepCFA} Bayesian posterior inference was approximated for all model parameters (e.g., via Gibbs sampling), which scales poorly. Here we employ Monte Carlo expectation maximization (MCEM) \citep{MCEM}, with a point estimate learned for the dictionary elements and the parameters of the classifier, allowing learning on large-scale data and fast testing.

Forms of stochastic pooling have been applied previously \citep{Lee09ICML,Zeiler13ICLR}.  \citet{Lee09ICML} defined stochastic pooling in the context of an energy-based Boltzmann machine, and \citet{Zeiler13ICLR} proposed stochastic pooling as a regularization technique. Here \emph{un}pooling is employed, yielding a top-down generative process. 

To impose supervision, we employ the Bayesian support vector machine (SVM)~\citep{polson11a}, which has been used for supervised dictionary learning \citep{Henao2014} (but not previously for deep learning). The proposed generative model is amenable to Bayesian analysis, and here the Bayesian SVM is learned simultaneously with the deep model. The models in~\citet{decaf,HeSSPnet,Zeiler14ECCV} do not train the SVM jointly, as we do -- instead, the SVM is trained separately using the learned CNN features (with CNN supervised learning implemented via softmax).

This paper makes several contributions: (\textit{i}) A new deep model is developed for images, based on convolutional dictionary learning; this model is a generative form of the earlier DN. (\textit{ii}) A new stochastic unpooling method is proposed, linking consecutive layers of the deep model. (\textit{iii}) An SVM is integrated with the top layer of the model, enabling max-margin supervision during training.  (\textit{iv}) The algorithm is implemented on a GPU, for large-scale learning and fast testing; we demonstrate state-of-the-art classification results on several benchmark datasets, and demonstrate scalability through analysis of the ImageNet dataset.

\section{Supervised Deep Deconvolutional Model \label{Sec:supDM}} 

\subsection{Single layer convolutional dictionary learning \label{Sec:cfa}}

Consider $N$ images $\{{\Xmat^{(n)}}\}_{n=1}^{N}$, with $\Xmat^{(n)}\in\mathbb{R}^{N_x \times N_y\times N_c}$, where $N_x$ and $N_y$ represent the number of pixels in each spatial dimension; $N_c=1$ for gray-scale images and $N_c=3$ for RGB images. 
We start by relating our model to optimization-based dictionary learning and DN, the work of \citep{Mairal08superviseddictionary,zhang10CVPR} and \citep{ciresan2012multi,Zeiler14ECCV,Zeiler10CVPR}, respectively. The motivations for and details of our model are elucidated by making connections to this previous work. Specifically, consider the optimization problem
{\small
	\begin{align}
	&\{\hat{\Dmat}^{(k)},\hat{\Smat}^{(n,k)}\} =\underset{}{\operatorname{argmin}} \left\{\sum_{n=1}^N \left\|\Xmat^{(n)}-\textstyle{ \sum_{k=1}^K\Dmat^{(k)} } * \Smat^{(n,k)}\right\|_F^2\right.\nonumber\\
	& \hspace{1.5cm}\left.+\lambda_1\sum_{k=1}^K \|\Dmat^{(k)}\|_F^2+\lambda_2\sum_{n=1}^N\sum_{k=1}^K\|\Smat^{(n,k)}\|_1\right\} \label{eq:opt}
	\end{align}
}
where $\ast$ is the 2D (spatial) convolution operator. Each $\Dmat^{(k)}\in\mathbb{R}^{n_x\times n_y\times N_c}$ and typically $n_x \ll N_x$, $n_y\ll N_y$. The spatially-dependent weights ${\bf S}^{(n,k)}$ are of size $(N_x-n_x +1)\times(N_y - n_y +1)$. Each of the $N_c$ layers of $\Dmat^{(k)}$ are spatially convolved with ${\bf S}^{(n,k)}$, and after summing over the $K$ dictionary elements, this manifests an approximation for each of the $N_c$ layers of $\Xmat^{(n)}$.

The form of (\ref{eq:opt}) is as in~\citet{Mairal08superviseddictionary}, with the $\ell_1$ norm on $\Smat^{(n,k)}$ imposing sparsity, and with the Frobenius norm on $\Dmat^{(k)}$ ($\ell_2$ in  \citet{Mairal08superviseddictionary}) imposing an expected-energy constraint on each dictionary element; in \citet{Mairal08superviseddictionary} convolution is not used, but otherwise the model is identical, and the computational methods developed in \citet{Mairal08superviseddictionary} may be applied. 

The form of (\ref{eq:opt}) motivates choices for the priors in the proposed generative model. Specifically, consider 
\begin{align}
\Xmat^{(n)}&=\sum_{k=1}^K\Dmat^{(k)} * ~\Smat^{(n,k)}+\Emat^{(n)},\label{eq:layer1}\\
\Emat^{(n)}&\sim\mathcal{N}(0,\gamma_e^{-1}\Imat),\qquad \Dmat^{(k)}\sim\mathcal{N}(0,\Imat)\label{eq:basic}
\end{align} 
with $S_{i,j}^{(n,k)}$ denoting element $(i,j)$ of $\Smat^{(n,k)}$, drawn $S_{i,j}^{(n,k)}\sim \mathsf{Laplace}(0,b)=\frac{1}{2b}\exp(-|S_{i,j}^{(n,k)}|/b)$. We have ``vectorized'' the matrices $\Emat^{(n)}$ and $\Dmat^{(k)}$ (from the standpoint of the distributions from which they are drawn), and $\Imat$ is an appropriately sized identity matrix. The maximum \textit{a posterior} (MAP) solution to (\ref{eq:basic}), with the Laplace prior imposed independently on each component of $\Smat^{(n,k)}$, corresponds to the optimization problem in (\ref{eq:opt}), and the hyperparameters $\gamma_e$ and $b$ play roles analogous to $\lambda_1$ and $\lambda_2$. 

The sparsity of $\Smat^{(n,k)}$ manifested in (\ref{eq:opt}) is a consequence of the geometry imposed by the $\ell_1$ operator; the MAP solution is sparse, but, with probability one, any draw from the Laplace prior on $\Smat^{(n,k)}$ is not sparse \citep{Volkan}. To impose sparsity on $\Smat^{(n,k)}$ within the generative process, we consider the spike-slab \citep{spikeslab} prior:
\begin{align}
S_{i,j}^{(n,k)}&\sim [z_{i,j}^{(n,k)} \mathcal{N}(0,\gamma_s^{-1})+(1-z_{i,j}^{(n,k)})\delta_0],\nonumber\\
z_{i,j}^{(n,k)}&\sim \mathsf{Bern}(\pi^{(n,k)}),\quad\pi^{(n,k)}\sim\mathsf{Beta}(a_0,b_0)\label{eq:spike}
\end{align}
where $z_{i,j}^{(n,k)}\in\{0,1\}$, $\delta_0$ is a unit point measure concentrated at zero, and $(a_0,b_0)$ are set to encourage that most $\pi^{(n,k)}$ are small \citep{Paisley09ICML}, \ie, $a_0=1/K$ and $b_0=1-1/K$. For parameters $\gamma_e$ and $\gamma_s$ we impose the priors $\gamma_s \sim \mathsf{Gamma}(a_s, b_s)$ and $\gamma_e \sim \mathsf{Gamma}(a_e, b_e)$, with hyperparameters $a_s=b_s=a_e=b_e=10^{-6}$ to impose diffuse priors \citep{RVM}.

\subsection{Generative Deep Model via Stochastic Unpooling\label{Sec:sto_pool}}
The model in (\ref{eq:layer1}) is motivated by the idea that each image $\Xmat^{(n)}$ may be represented in terms of convolutional dictionary elements $\Dmat^{(k)}$ that are shared across all $N$ images. In the proposed deep model, we similarly are motivated by the idea that the feature maps $\Smat^{(n,k)}$ may also be represented in terms of convolutions of (distinct) dictionary elements. Consider a two-layer model, with 
 \begin{align}
{{\bf X}^{(n,2)}}&= \sum_{k_{2}=1}^{K_{2}} {\bf D}^{(k_{2}, 2)} * \Smat^{(n,k_2,2)} \label{Eq:x_2}\\
\Smat^{(n,k_1,1)}&\sim\mathsf{unpool}({\bf X}^{(n,k_1,2)})~,~~~k_1=1,\ldots,K_1\label{eq:unpool}\\
	\Xmat^{(n,1)}&= \sum_{k_{1}=1}^{K_{1}} {\bf D}^{(k_{1}, 1)} * \Smat^{(n,k_1,1)}+ {\bf E}^{(n)} \label{Eq:x_1}
\end{align}
where $\Xmat^{(n,1)}=\Xmat^{(n)}$. Dictionary elements $\Dmat^{(k_1,1)}$ replace $\Dmat^{(k)}$ in (\ref{eq:layer1}), for representation of $\Xmat^{(n)}$. The weights $\Smat^{(n,k_1,1)}$ are connected to $\Xmat^{(n,2)}$ via the stochastic operation $\mathsf{unpool}(\cdot)$, detailed below. Motivated as discussed above, $\Xmat^{(n,2)}$ is represented in terms of convolutions with second-layer dictionary elements $\Dmat^{(k_2,2)}$. The forms of the priors on $\Dmat^{(k_1,1)}$ and $\Dmat^{(k_2,2)}$ are as above for $\Dmat^{(k)}$, and the prior on $\Emat^{(n)}$ is unchanged. 

The tensor $\Xmat^{(n,2)}\in\mathbb{R}^{N_x^{(2)}\times N_y^{(2)}\times K_1}$ with layer/slice $k_1$ denoted by the matrix $\Xmat^{(n,k_1,2)}\in\mathbb{R}^{N_x^{(2)}\times N_y^{(2)}}$ ($k_1\in\{1,\dots,K_1\}$). Matrix $\Xmat^{(n,k_1,2)}$ is a {\em pooled} version of $\Smat^{(n,k_1,1)}$. Specifically, $\Smat^{(n,k_1,1)}$ is partitioned into contiguous spatial pooling blocks, each pooling block of dimension $p_x^{(1)}\times p_y^{(1)}$, with $N_x^{(2)}=N_x/p_x^{(1)}$ and $N_y^{(2)}=N_y/p_y^{(1)}$ (assumed to be integers). Each pooling block of $\Smat^{(n,k_1,1)}$ is all-zeros except one non-zero element, with the non-zero element defined in $\Xmat^{(n,k_1,2)}$. Specifically, element $(i,j)$ of $\Xmat^{(n,k_1,2)}$, denoted $X^{(n,k_1,2)}_{i,j}$, is mapped to pooling block $(i,j)$ in $\Smat^{(n,k_1,1)}$, denoted $\Smat^{(n,k_1,1)}_{i,j}$. 

\begin{figure}[tb]
	\centering
	\vspace{-1pt}
	\includegraphics[scale=0.6]{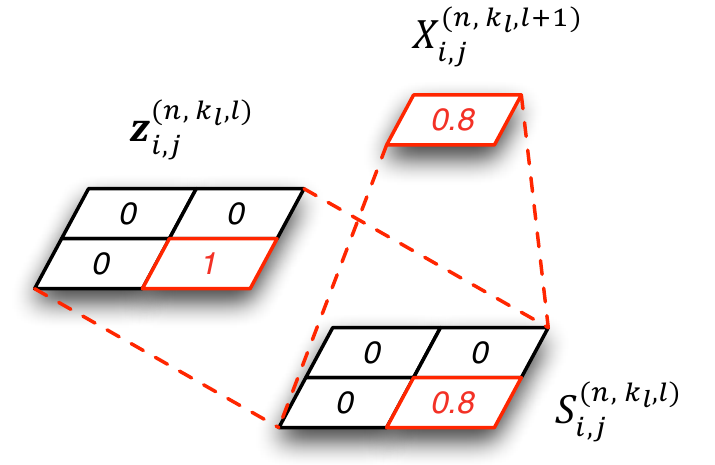}
	\caption{\small{Demonstration of the stochastic unpooling; one $2\times 2$ pooling block is depicted.}}
	\label{fig:sto_pool}
\end{figure}

Let $\zv_{i,j}^{(n,k_1,1)}\in\{0,1\}^{p_x^{(1)}p_y^{(1)}}$ be a vector of all zeros, and a single one, and the location of the non-zero element of $\zv_{i,j}^{(n,k_1,1)}$ identifies the location of the single non-zero element of pooling block $\Smat^{(n,k_1,1)}_{i,j}$ which is set as $X^{(n,k_1,2)}_{i,j}$. The function $\mathsf{unpool}(\cdot)$ is a stochastic operation that defines $\zv_{i,j}^{(n,k_1,1)}$, and hence the way $\Xmat^{(n,k_1,2)}$ is {\em unpooled} to constitute the {\em sparse} $\Smat^{(n,k_1,2)}$. We impose
\begin{align}
\zv_{i,j}^{(n,k_1,1)}&\sim \mathsf{Mult}(1,\thetav^{(n,k_1,1)})\\
\thetav^{(n,k_1,1)}&\sim\mathsf{Dir}(1/(p_x^{(1)}p_y^{(1)}))\label{eq:unpool}
\end{align}
where $\mathsf{Dir}(\cdot)$ denotes the symmetric Dirichlet distribution; the Dirichlet distribution has a {\em set} of parameters, and here they are all equal to the value indicated in $\mathsf{Dir}(\cdot)$. 

When introducing the above two-layer model, $\Smat^{(n,k_2,2)}$ is drawn from a spike-slab prior, as in (\ref{eq:spike}). However, we may extend this to a three-layer model, with pooling blocks defined in $\Smat^{(n,k_2,2)}$. A convolutional dictionary representation is similarly constituted for $\Xmat^{(n,3)}$, and this is stochastically unpooled to generate $\Smat^{(n,k_2,2)}$. This may continued for $L$ layers, where the hierarchical convolutional dictionary learning learns multi-scale structure in the weights on the dictionary elements. At the top layer in the $L$-layer model, the weights $\Smat^{(n,k_L,L)}$ are drawn from a spike-slab prior of the form in (\ref{eq:spike}).

Consider an $L$-layer model, and assume that $\{\{\Dmat^{(k_l,l)}\}_{k_l=1}^{K_l}\}_{l=1}^{L}$ have been learned/specified. An image is generated by starting at the top, drawing  $\{\Smat^{(n,k_L,L)}\}_{k_L=1}^{K_L}$ from a spike-slab model. Then $\{\Smat^{(n,k_{L-1},L-1)}\}_{k_{L-1}=1}^{K_{L-1}}$ are constituted by convolving $\{\Smat^{(n,k_L,L)}\}_{k_L=1}^{K_L}$ with $\{\Dmat^{(k_L,L)}\}_{k_L=1}^{K_L}$, summing over the $K_L$ dictionary elements, and then performing stochastic unpooling. This process of convolution and stochastic unpooling proceeds for $K_L$ layers, ultimately yielding $\sum_{k_{1}=1}^{K_{1}} {\bf D}^{(k_{1}, 1)} * \Smat^{(n,k_1,1)}$ at the bottom (first) layer. With the added stochastic residual $\Emat^{(n)}$, the image $\Xmat^{(n)}$ is specified.

We note an implementation detail that has been found useful in experiments. In (\ref{eq:unpool}), the unpooling was performed such that each pooling block in $\Smat^{(n,k_l,l)}$ has a single non-zero element, with the non-zero element defined in $\Xmat^{(n,k_l,l+1)}$. The unpooling for block $(i,j)$ was specified by the $p_x^{(l)}p_y^{(l)}$-dimensional $\zv_{i,j}^{(n,k_l,l)}$ vector of all-zeros and a single one. In our slightly modified implementation, we have considered a $(p_x^{(l)}p_y^{(l)}+1)$-dimensional $\zv_{i,j}^{(n,k_l,l)}$, which is again all zeros with a single one, and $\thetav^{(n,k_l,l)}$ is also $p_x^{(l)}p_y^{(l)}+1$ dimensional. If the single one in $\zv_{i,j}^{(n,k_l,l)}$ is located  among the first $p_x^{(l)}p_y^{(l)}$ elements of $\zv_{i,j}^{(n,k_l,l)}$, then the location of this non-zero element identifies the location of the single non-zero element in the $(i,j)$ pooling block, as before. However, if the non-zero element of $\zv_{i,j}^{(n,k_l,l)}$ is in position $p_x^{(l)}p_y^{(l)}+1$, then {\em all} elements of pooling block $(i,j)$ are set to zero. This imposes further sparsity on the feature maps and, as demonstrated in the Supplementary Material (SM), it yields a model in which the elements of the feature map that are relatively small are encouraged to be zero. This turning off of dictionary elements with small weights is analogous to {\em dropout} \citep{dropout}, which has been used in CNN, and in our model it also has been found to yield slightly better classification performance.
\vspace{-2mm}
\subsection{Supervision via Bayesian SVMs} 
\label{Sec:sDcFA}
\vspace{-1mm}
Assume that a label $\ell_n\in\{1,\dots,C\}$ is associated with each of the $N$ images, so that the training set may be denoted $\{(\Xmat^{(n)},\ell_n)\}_{n=1}^{N}$. 
We wish to learn a classifier that maps the {\em top-layer} dictionary weights $\Smat^{(n,L)}=\{\Smat^{(n,k_l,L)}\}_{k_l=1}^{K_L}$ to an associated label $\ell_n$. The $\Smat^{(n,L)}$ are ``unfolded'' into the vector $\sv_n$. We desire the classifier mapping $\sv_n\rightarrow \ell_n$ and our goal is to learn the dictionary and classifier \emph{jointly}. 

We design $C$ one-versus-all binary SVM classifiers. For each of these classifiers, the problem may be posed as training with $\{\sv_n,y_n^{(\ell)}\}_{n=1}^{N}$, where $\sv_n$ are the top-layer dictionary weights, as discussed above, and $y_n^{(\ell)}\in\{-1,1\}$ (a bias term is also appended to each $\sv_n$, as is typical of SVMs). If $\ell_n=\ell\in\{1,\dots,C\}$ then $y_n^{(\ell)}=1$, and $y_n^{(\ell)}=-1$ otherwise; the indicator $\ell$ specifies which of the $C$ binary SVMs is under consideration. For notational simplicity, we omit the superscript $(\ell)$ for the remainder of the section, and consider the Bayesian SVM for one of the binary learning tasks, with labeled data $\{\sv_n,y_n\}_{n=1}^{N}$. In practice, $C$ such binary classifiers are learned jointly, and the value of $y_n\in\{1,-1\}$ depends on which one-versus-all classifier is being specified.

Given a feature vector $\sv$, the goal of the SVM is to find an $f(\sv )$ that minimizes the objective function
\beq\label{eq:svm}
\textstyle{\gamma\sum_{n=1}^N \max (1-y_nf(\sv_n),0)} + R(f(\sv)),
\eeq
where $\max(1-y_nf(\sv_n),0)$ is the hinge loss, $R(f(\sv))$ is a regularization term that controls the complexity of $f(\sv)$, and $\gamma$ is a tuning parameter controlling the trade-off between error penalization and the complexity of the classification function. The decision boundary is defined as $\{\sv :f(\sv )=0\}$ and $\mbox{sign}(f(\sv))$ is the decision rule, classifying $\sv$ as either $-1$ or $1$~\citep{vapnik95statLearning}.

Recently, \citet{polson11a} showed that for the linear classifier $f(\sv)=\betav^T\sv$, minimizing \eqref{eq:svm} is equivalent to estimating the mode of the pseudo-posterior of $\betav$
\begin{align}\label{eq:pos_beta}
\textstyle p(\betav|\Smat,\yv,\gamma) \propto \prod_{n=1}^N \mathcal{L}(y_n|\sv_n,\betav,\gamma)p(\betav|\cdot) \,,
\end{align}
where $\yv=[y_1 \ \ldots \ y_N]^T$, $\Smat=[\sv_1 \ \ldots \ \sv_N]$, $\mathcal{L}(y_n|\sv_n,\betav,\gamma)$ is the pseudo-likelihood function, and $p(\betav|\cdot)$ is the prior distribution for the vector of coefficients $\betav$. Choosing $\betav$ to maximize the log of \eqref{eq:pos_beta} corresponds to \eqref{eq:svm}, where the prior is associated with $R(f(\sv))$. ~\citet{polson11a} showed that $\mathcal{L}(y_n|\sv_n,\betav,\gamma)$ admits a location-scale mixture of normals representation by introducing latent variables $\lambda_n$, such that
\beqs
& \mathcal{L}(y_n|\sv_n,\betav,\gamma) = e^{-2\gamma\max(1-y_n\betav^T\sv_n,0)} \nonumber\\
&\qquad  =  \int_0^\infty \frac{\sqrt{\gamma}}{\sqrt{2\pi\lambda_n}}\exp\left(-\frac{(1+\lambda_n-y_n\betav^T\sv_n)^2}{2\gamma^{-1}\lambda_n}\right) d\lambda_n. \quad\label{eq:lik_mix}
\eeqs
Note that the exponential in (\ref{eq:lik_mix}) is Gaussian wrt $\betav$. As described in \citet{polson11a}, this encourages data augmentation for variable $\lambda_n$ ($\lambda_n$ is treated as a new random variable), which permits efficient Bayesian inference (see \citet{polson11a,Henao2014} for details).
One of the benefits of a Bayesian formulation for SVMs is that we can flexibly specify the behavior of $\betav$ while being able to adaptively regularize it by specifying a prior $p(\gamma)$ as well. 

We impose shrinkage (near sparsity)~\citep{Polson10shrinkglobally} on $\betav$ using the Laplace distribution; letting $\beta_i$ denote $i^{\rm th}$ element of $\betav$, we impose 
\vspace{-2mm}
\begin{equation}\label{eq:beta}
\small
\beta_i\sim {\cal N}(0,\omega_i), \quad \omega_i \sim \mathsf{Exp}(\kappa),\quad \kappa\sim\mathsf{Gamma}(a_{\kappa}, b_{\kappa}),
\end{equation}
and similar to $\kappa$ and $\lambda_n$, a diffuse Gamma prior is imposed on $\gamma$.


For the generative process of the overall model, activation weights $\sv_n$ are drawn at layer $L$, as discussed in Sec. \ref{Sec:sto_pool}. These weights then go into the $C$-class SVM, and from that a class label is manifested. 
Specifically, each SVM learns a linear function of $\{\betav_{\ell}^\top\sv\}_{\ell=1}^C$, and for a given data $\sv$, its class label is defined by~\citep{yang09CVPR}: \beq \ell_n = \underset{\ell}{\operatorname{argmax}} ~~\betav_{\ell}^\top\sv_n.\eeq
The set of vectors $\{\betav_\ell\}_{\ell=1}^{C}$, connecting the top-layer features $\sv$ to the classifier, play a role analogous to the fully-connected layer in the softmax-based CNN, but here we constitute supervision via the max-margin SVM.
Hence, the proposed model is a generative construction for both the labels and the images. 
\vspace{-3mm}
\section{Model Training}\label{Sec:train}
\vspace{-3mm}
The previous section described a supervised deep generative model for images, based on deep convolutional dictionary learning, stochastic unpooling, and the Bayesian SVM. The conditional posterior distribution for each model parameter can be written in closed form, assuming the other model parameters are fixed (see the SM). For relatively small datasets we can therefore employ a Gibbs sampler for both training and deconvolutional inference, yielding an approximation to the posterior distribution on all parameters. Large-scale datasets prohibit the application of standard Gibbs sampling. For large data we use stochastic MCEM \citep{MCEM} to find a maximum \textit{a posterior} (MAP) estimate of the model parameters.

We consolidate the ``local'' model parameters (latent data-sample-specific variables) as\\[-0.5em]
\[\Phimat_n=\big(\{\zv^{(n,l)}\}_{l=1}^L,{\Smat}^{(n,L)},{\gammav}_s^{(n)},{\Emat}^{(n)},\{{\lambda}_n^{(\ell)}\}_{\ell=1}^C\big),\]\\[-1.25em]
the ``global'' parameters (shared across all data) as \( \Psimat=\big(\{{\Dmat}^{(l)}\}_{l=1}^L,\betav\big),\) and the data as \(\Ymat_n=(\Xmat^{(n)},\ell_n).\)
We desire a MAP estimator
\begin{align}
\Psimat_{\text{MAP}}=\underset{\Psimat}{\operatorname{argmax}}\ \sum_n \ln p({\Psimat}|\Ymat_n),
\label{eq:map}
\end{align}
\vspace{-1mm}
which can be interpreted as an EM problem:\\[-2em]
\paragraph{E-step:}\hspace{-6pt}Perform an expectation with respect to the local variables, using \(p(\Phimat_n|\Ymat_n, \Psimat_{t-1})\,\forall n\), where $\Psimat_{t-1}$ is the estimate of the global parameters from iteration $(t-1)$. \\[-2.25em]
\paragraph{M-step:}\hspace{-6pt}Maximize~~\(\ln p(\Psimat) + \sum_n \mathbb{E}_{\Phimat_n}[\ln p(\Ymat_n|\Phimat_n, \Psimat)]\) with respect to $\Psimat$.\\[0.5em]
We approximate the expectation via Monte Carlo sampling, which gives
{\small
\begin{align}
Q(\Psimat|\Psimat_{t-1}) =\ln p(\Psimat) + \frac{1}{N_s}\sum_{s=1}^{N_s} \sum_n\ln p(\Ymat_n|\Phimat_n^s, \Psimat),
\end{align}}
where $\Phimat_n^s$ is a sample from the full conditional posterior distribution, and $N_s$ is the number of samples; we seek to maximize $Q(\Psimat|\Psimat_{t-1})$ wrt $\Psimat$, constituting $\Psimat_t$. Recall from above that each of the conditional distributions in a Gibbs sampler of the model is analytic; this allows convenient sampling of local parameters, conditioned on specified global parameters $\Psimat_{t-1}$, and therefore the aforementioned sampling is implemented efficiently (using mini-batches of data, where $\mathcal{I}_t \subset\{1,\dots,N\}$ identifies the stochastically defined subset of data in mini-batch $t$). An approximation to the M-step is implemented via stochastic gradient descent (SGD). The stochastic MCEM gradient at iteration $t$ is 
\beq
\nabla_{\Psimat}Q = \nabla_{\Psimat}\ln p(\Psimat) + \frac{1}{N_s}\sum_{s=1}^{N_s} \sum_{n\in \mathcal{I}_t} \nabla_{\Psimat}\ln p(\Ymat_n|\Phimat_n^s, \Psimat).
\label{eq:gradient}
\eeq
We solve (\ref{eq:map}) using RMSprop \citep{rms15,rmsprop} with the gradient approximation in (\ref{eq:gradient}).

In the learning phase, the MCEM method is used to learn a point estimate for the global parameters $\Psimat$. During testing, we follow the same MCEM setup with \(\Phimat^\text{test}=\big(\{\zv^{(*,l)}\}_{l=1}^{L-1},{\gammav}_s^{(*)},{\Emat}^{(*)}\big)\) ,\(\quad\Psimat^{\text{test}}=\Smat^{(*,L)},\) when given a new image $\Xmat^{*}$. We find a MAP estimator:
\begin{align}\label{Eq:test}
\Psimat^{\text{test}}_\text{MAP}=\underset{\Psimat^{\text{test}}}{\operatorname{argmax}}\ \ln p(\Psimat^{\text{test}}|\Xmat^{*},\Dmat),
\end{align}
using MCEM (gradient wrt $\Psimat^{\text{test}}$). In this form of the MCEM, all data-dependent latent variables $\Phimat^\text{test}$ are integrated (summed) out in the expectation, except for the top-layer feature map $\Psimat^{\text{test}}$, for which the gradient descent M step yields a point estimate. The top-layer features are then sent to the trained SVM to predict the label. Details for training and inference are provided in the SM.
\vspace{-2mm}
\section{Experimental Results \label{Sec:Experiment}}
\vspace{-2mm}
We present results for the MNIST, CIFAR-10 \& 100, Caltech 101 \& 256 and ImageNet 2012 datasets. The same hyperparameter settings (discussed at the end of Section \ref{Sec:cfa}) were used in all experiments; no tuning was required between datasets. 

For the first five (small/modest-sized) datasets, the model is learned via Gibbs sampling. We found that it is effective to use layer-wise pretraining as employed in some deep generative models~\citep{Erhan10Pretrain,Hinton06Science}. The pretraining is performed sequentially from the bottom layer (touching the data), to the top layer, in an unsupervised manner. Details on the layerwise pretraining are discussed in the SM. 
In the pretraining step, we average 500 collection samples, to obtain parameter values (\eg, dictionary elements) after first discarding 1000 burn-in samples. 
Following pre-training, we refine the entire model jointly using the complete set of Gibbs conditional distributions. 1000 burn-in iterations are performed followed by 500 collection draws, retaining one of every 50 iterations. During testing, the predictions are based on averaging the decision values of the collected samples.

For each of these first five datasets, we show three classification results, using part of or all of our model (to illustrate the role of each component): 
1) {\em Pretraining only}: this model (in an unsupervised manner) is used to extract features and the futures are sent to a separate linear SVM, yielding a 2-step procedure.
2) {\em Unsupervised model}: this model includes the deep generative developed in Sec.~\ref{Sec:sto_pool}, but is also trained in an unsupervised manner (this is the unsupervised model after refinement). The features extracted by this model are sent to a separate linear SVM, and therefore this is also a 2-step procedure.
3) {\em Supervised model}: this is the complete refined supervised model developed in Sec.~\ref{Sec:sto_pool} and Sec.~\ref{Sec:sDcFA}.

ImageNet 2012 is used to assess the scalability of our model to large datasets.  In this case, we learn the supervised model initialized from the priors (without layerwise pretraining). The proposed online learning method, MCEM, based on RMSProp~\citep{rms15,rmsprop}, is developed  for both training and inference with mini-batch size 256 and decay rate 0.95. Our implementation of MCEM learning is based on the publicly available CUDA C++ Caffe toolbox (August 2015 branch) \citep{caffe}, but contains significant modifications for our model. 
Our model takes around one week to train on ImageNet 2012 using a nVidia GeForce GTX TITAN X GPU with 12GB memory. Testing for the validation set of ImageNet 2012 (50K images) takes less than 12 minutes.
In the subsequent tables providing classification results, the best results achieved by our model are bold.

\vspace{-3mm}
\subsection{MNIST}
\vspace{-2mm}
The MNIST data (\url{http://yann.lecun.com/exdb/
	mnist/}) has 60,000 training and 10,000
testing images, each $28\times28$, for digits 0 through 9.
A two-layer model is used with dictionary element size $8\times 8$ and $6\times 6$ at the first and second layer, respectively.
The pooling size is $3\times 3$ ($p_x=p_y=3$) and the number of dictionary elements at layers 1 and 2 are $K_1=39$ and $K_2=117$, respectively. 
These numbers of dictionary elements are obtained by setting the initial dictionary number to a relatively large value ($K_1=50$ and $K_2=200$) in the pretraining step and discarding infrequently used elements by counting the corresponding binary indicator $\zv$ -- effectively \emph{inferring} the number of needed dictionary elements, as in~\citet{Chen2011ICML,Chen13deepCFA}.
\begin{table}
	\vspace{-4mm}
	\caption{\small{Classification Error ($\%$) for MNIST}}
	\vspace{-3mm}
	\centering
	\small
	\begin{tabular}{c|c}
		\hline
		\bf{Method} & \bf{Test error} \\
		\hline
		{2-layer convnet \citep{Jarrett09ICCV}} &  0.53 \\
		Our pretrained model + SVM 
		& 1.42 \\
		Our unsupervised model + SVM & 0.52 \\
		Our supervised model & {\bf 0.37} \\
		{6-layer convnet \citep{Ciresan11IJCAI}}
		& 0.35 \\
		MCDNN~\citep{ciresan2012multi} & 0.23\\	
		\hline
	\end{tabular}
	\label{Table:Error_MNIST}
	\vspace{-2mm}
\end{table}

Table~\ref{Table:Error_MNIST} summarizes the classification results for MNIST. Our 2-layer supervised model outperforms most other modern approaches. 
The methods that outperforms ours are the complicated (6-layer) ConvNet model with elastic disortions~\citep{Ciresan11IJCAI} and the MCDNN, which combines several deep convolutional neural networks~\citep{ciresan2012multi}. Specifically,~\citep{ciresan2012multi} used a committee of 35 convolutional networks, width normalization, and elastic distortions of the data; \citep{Ciresan11IJCAI} used elastic distortions and a single convolutional neural network to achieve the similar error as our approach.

\begin{figure}[tb]
	\centering
	\vspace{-1mm}
	\includegraphics[scale=0.42]{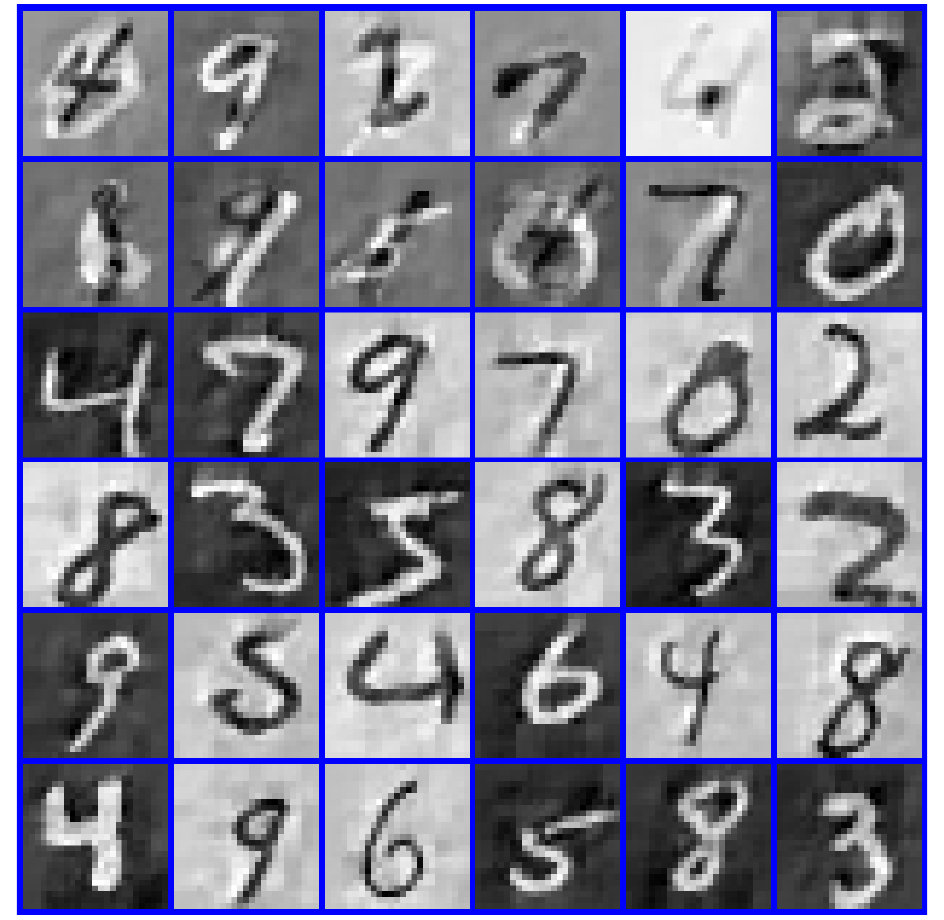}
	\includegraphics[scale=0.42]{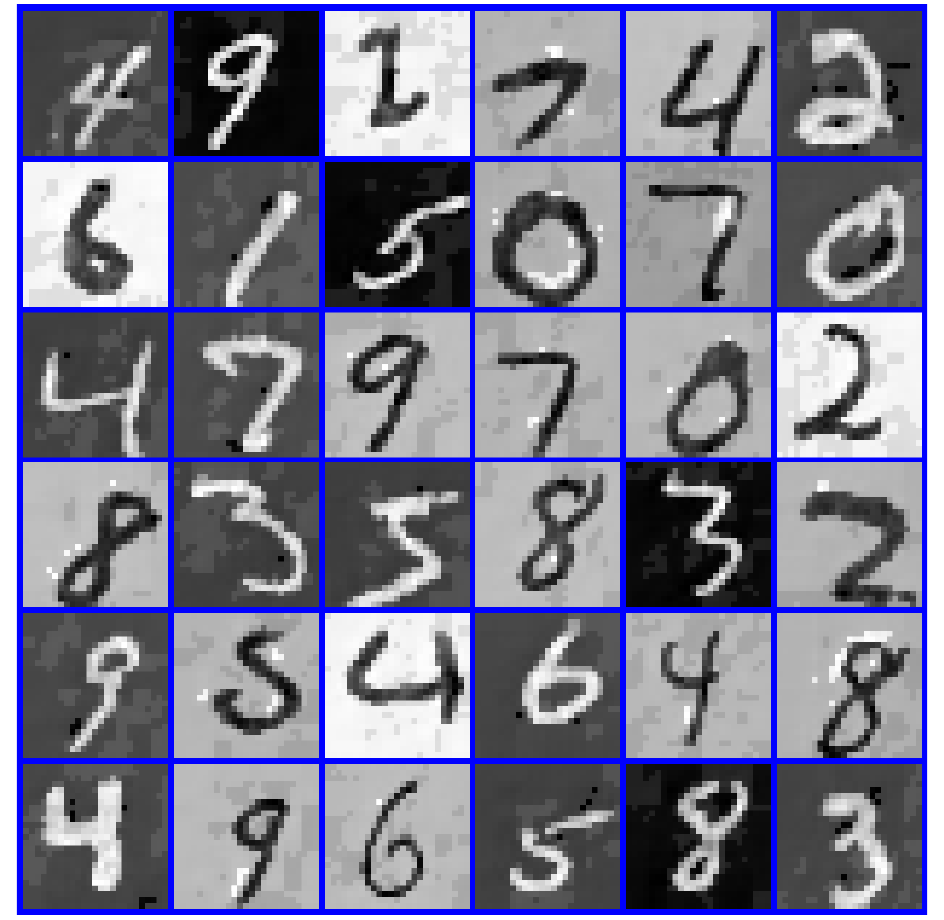}
	\vspace{-3mm}
	\caption{\small{Selected layer 2 (top-layer) dictionary elements of MNIST learned by the unsupervised model (left) and the supervised model (right).}}
	\vspace{-5mm}
	\label{fig:MNISTsuper_unsper}
\end{figure}

To further examine the performance of the proposed model, we plot a selection of top-layer dictionary elements (projected through the generative process down to the data plane) learned by our {supervised model, on the right of Fig.~\ref{fig:MNISTsuper_unsper}, and on the left we show the corresponding elements inferred by our unsupervised model.} It can be seen that the elements inferred by the supervised model are clearer (``unique" to a single number), whereas the elements learned by the unsupervised model are blurry (combinations of multiple numbers). Similar results were reported in~\citet{Erhan10Pretrain}. 

\begin{figure}[tb]
	\centering
	\includegraphics[trim = 0.5cm 0 0 10.61cm, clip, width=0.5\textwidth]
	 {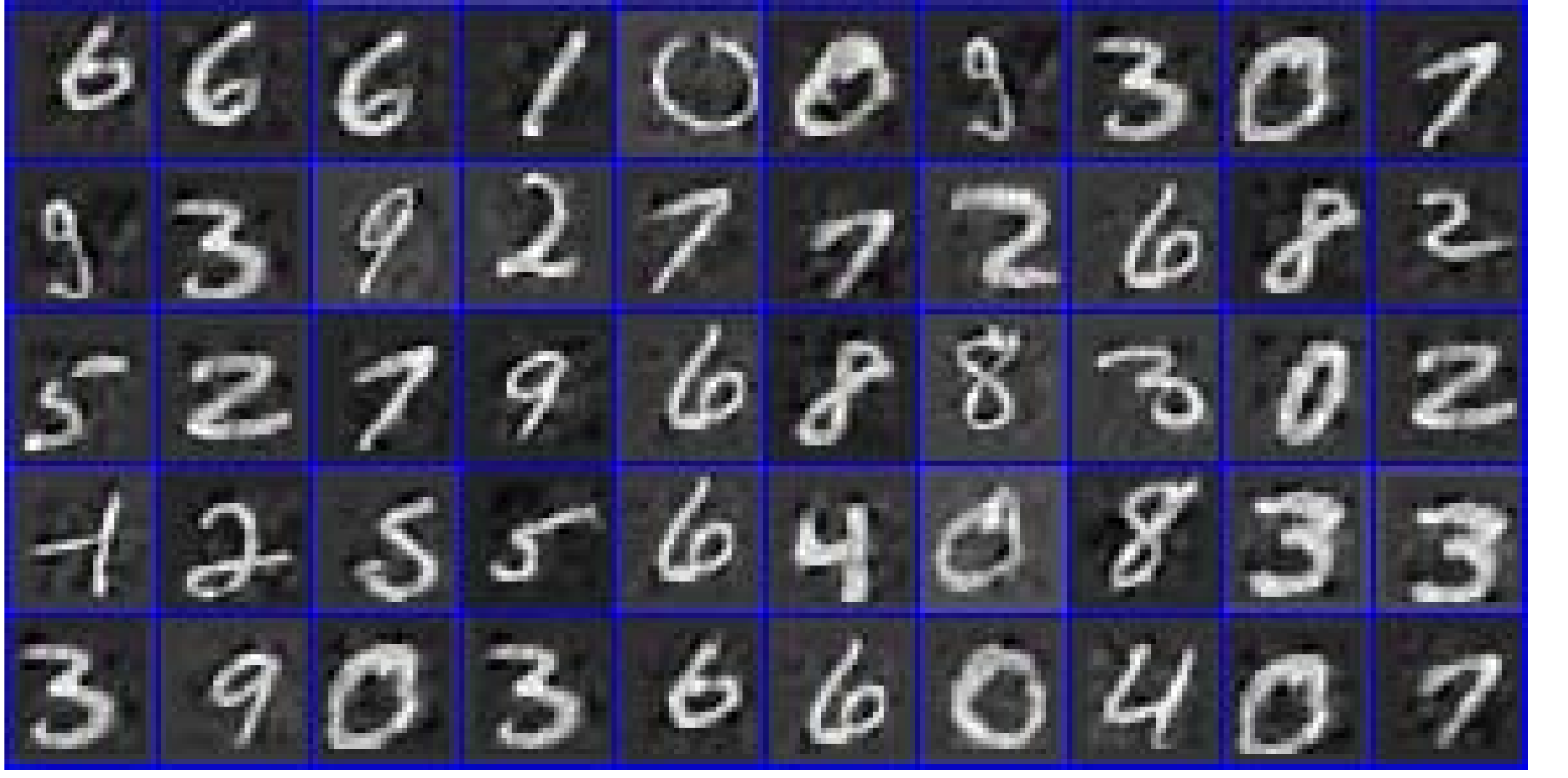}
	 \vspace{-7mm}
	 \caption{\small{Generated images using random dictionary weights.}}
	 \label{fig:MNIST_Generate}
	 \vspace{-2mm}
\end{figure}

\begin{figure}[htb]
	\centering
	\includegraphics[width=0.48\textwidth]{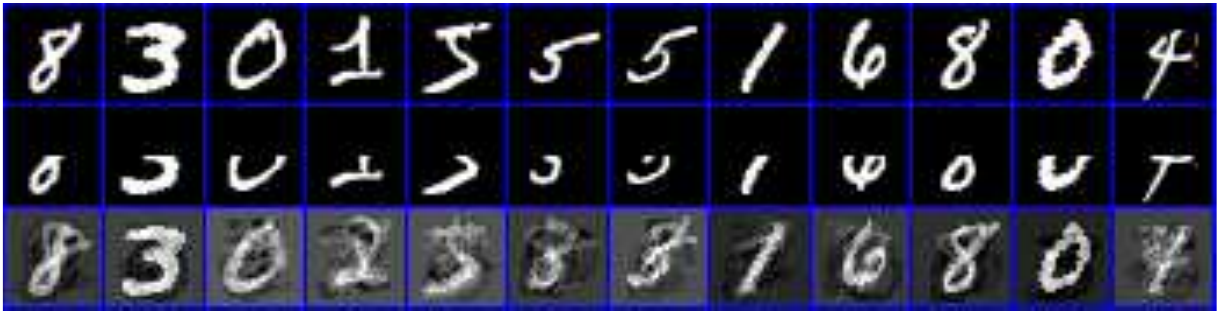}\\
	\includegraphics[width=0.48\textwidth]{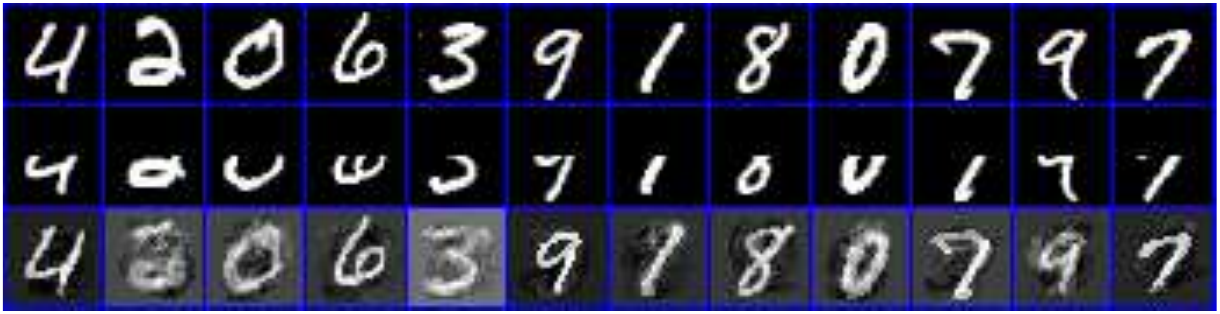}
	\vspace{-6mm}
	\caption{\small{Missing data interpolation of digits. For each subfigure: (top) Original data, (middle) Observed data,  (bottom) Reconstruction.}}
	\vspace{-5mm}
	\label{fig:MNIST_missing}
\end{figure}

Since our model is generative, using it to generate digits after training on MNIST is straightforward, and some examples are shown in Fig.~\ref{fig:MNIST_Generate} (based on random draws of the top-layer weights). 
We also demonstrate the ability of the model to predict missing data (generative nature of the model); reconstructions are shown in Fig. \ref{fig:MNIST_missing}. 
More results are provided in the SM.


\vspace{-3mm}
\subsection{CIFAR-10 \& 100}
\vspace{-2mm}
The CIFAR-10 dataset \citep{krizhevsky2009learning} is composed of 10 classes of natural $32\times 32$ RGB images with 50000 images for training and 10000 images for testing. We apply the same preprocessing technique of global contrast normalization and ZCA whitening as used in the Maxout network \citep{goodfellow2013maxout}.
A three-layer model is used with dictionary element size $5\times 5$, $5\times 5$, $4\times 4$ at the first, second and third layer.
The pooling sizes are both  $2\times 2$  and the numbers of dictionary elements for each layer are $K_1=48$, $K_2=128$ and $K_3=128$. 
If we {\em augment} the data by translation and horizontal flipping as used in other models \citep{goodfellow2013maxout}, we achieve $8.27\%$ error. Our result is competitive with the state-of-art, which integrates supervision on every hidden layer~\citep{lee2014deeply}. In constrast, we only impose supervision at the top layer.
Table~\ref{Table:Error_CIFAR10} summarizes the classification accuracy of our models and some related models.
\begin{table}[tp!]
	\caption{\small{Classification Error ($\%$) for CIFAR-10}}
	\vspace{-3mm}
	\centering
	\begin{small}
		\begin{tabular}{c|c}
			\hline
			\bf{Method} & \bf{Test error} \\
			\hline
			\textit{Without Data Augmentation}\\
			\hline
			Maxout~\citep{goodfellow2013maxout} & 11.68\\
			Network in Network~\citep{NIN} & 10.41\\	
			Our pretrained + SVM  & 22.43\\
			Our unsupervised + SVM & 14.75\\
			Our supervised model	& 10.39 \\
			Deeply-Supervised Nets~\citep{lee2014deeply} & 9.69\\
			\hline
			\textit{With Data Augmentation}\\
			\hline
			Maxout  ~\citep{goodfellow2013maxout} 	& 9.38\\
			Network in Network~\citep{NIN} & 8.81\\	
			Our pretrained + SVM  & 20.62\\
			Our unsupervised + SVM & 10.22\\
			Our supervised  & {\bf{8.27}} \\
			Deeply-Supervised Nets~\citep{lee2014deeply} & 7.97\\
			\hline
		\end{tabular}
	\end{small}
	\label{Table:Error_CIFAR10}
		\vspace{-4mm}
\end{table}

The CIFAR-100 dataset \citep{krizhevsky2009learning} is the same as CIFAR-10 in size and format, except it contains 100 classes. We use the same settings as in the CIFAR-10. Table~\ref{Table:Error_CIFAR100} summarizes the classification accuracy of our model and some related models. It can be seen that our results ($34.62\%$) are also very close to the state-of-the-art: ($34.57\%$) in \citet{lee2014deeply}. 

\begin{table}[htbp!]
	\vspace{-3mm}
	\caption{\small{Classification Error ($\%$) for CIFAR-100}}
	\vspace{-3mm}
	\centering
	\small
	\begin{tabular}{c|c}
		\hline
		\bf{Method} & \bf{Test error} \\
		\hline 
		Maxout~\citep{goodfellow2013maxout} 
		& 38.57\\
		Network in Network~\citep{NIN} & {35.68}\\
		Our pretrained + SVM (2 step)& 77.25  \\
		Our unsupervised + SVM (2 step)& 42.26\\
		Our supervised model  & {\bf 34.62} \\	
		Deeply-Supervised Nets~\citep{lee2014deeply} & 34.57\\
		\hline
	\end{tabular}
	\label{Table:Error_CIFAR100}
	\vspace{-3mm}
\end{table}

\vspace{-1mm}
\subsection{Caltech 101 \& 256}
\vspace{-3mm}
To balance speed and performance, we resize the images of Caltech 101 and Caltech 256 to $128\times 128$, followed by local contrast normalization \citep{Jarrett09ICCV}. A three layer model is adopted.
The dictionary element sizes are set to $7\times 7$, $5\times 5$ and $5\times 5$, and the size of the pooling regions are $4\times 4$ (layer 1 to layer 2) and $2\times 2$ (layer 2 to layer 3). 

The dictionary sizes for each layer are set to $K_1=48$, $K_2=84$ and $K_3=84$ for Caltech 101, and  $K_1=48$, $K_2=128$ and $K_3=128$ for Caltech 256. Tables~\ref{Table:caltech101} and~\ref{Table:caltech256} summarize the classification accuracy of our model and some related models. Using only the data inside Caltech 101 and Caltech 256 (without using other datasets) for training, our results ($87.82\%,\,66.4\%$) exceed the previous state-of-art results ($83\%,\,58\%$) by a substantial margin ($4\%,\,12.4\%$), which are the best results obtained by models without using deep convolutional models (using hand-crafted features).

As a baseline, we implemented the neural network consisting of three convolutional layers and two fully-connected layers with a final softmax classifier. The architecture of three convolutional layers is the same as our model. The fully-connected layers have 1024 neurons each. The results of neural network trained with dropout~\citep{dropout}, after carefully parameter tuning, are also shown in Tables~\ref{Table:caltech101} and~\ref{Table:caltech256}. 
\begin{table}[tp!]
	\small
	\vspace{-1mm}
	\caption{\small{Classification Accuracy ($\%$) for Caltech 101}}
	\vspace{-2mm}
	\centering
	\begin{tabular}{c|c c}
		\hline
		\textbf{Training images per class} & \bf{15} & \bf{30}\\
		\hline 
		\textit{Without ImageNet Pretrain}\\
		\hline
		5-layer Convnet~\citep{Zeiler14ECCV}  & 22.8 & 46.5 \\
		HBP-CFA~\citep{Chen13deepCFA} & 58 & 65.7\\
		R-KSVD~\citep{li2013reference} &  79  & 83 \\
		3-layer Convnet      & 62.3 & 72.4\\
		Our pretrained + SVM (2 step)& 43.24 & 53.57  \\
		Our unsupervised + SVM (2 step)& 70.47 & 80.39 \\
		Our supervised model &  75.37 & 87.82 \\
		\hline 
		\textit{With ImageNet Pretrain}\\
		\hline
		5-layer Convnet~\citep{Zeiler14ECCV}& 83.8  & 86.5\\
		5-layer Convnet~\citep{return}& -  & 88.35\\
		Our supervised model &  {\bf 89.1} & {\bf 93.15} \\
		SPP-net~\citep{HeSSPnet} & - & 94.11 \\
		\hline
	\end{tabular}
	\label{Table:caltech101}
	\vspace{-3mm}
\end{table}
\begin{table}[btbp!]
	\small
	\caption{\small{Classification Accuracy ($\%$) for Caltech 256} }
	\vspace{-3mm}
	\centering
	\begin{tabular}{c|cc}
		\hline
		\textbf{Training images per class} & \bf{15}  & \bf{60}\\
		\hline 
		\textit{Without ImageNet Pretrain}\\
		\hline
		5-layer Convnet~\citep{Zeiler14ECCV}  & 9.0  & 38.8\\ 
		Mu-SC~\citep{BoCVPR2013}  & 42.7 & 58 \\
		3-layer Convet & 46.1  & 60.1\\
		Our pretrained +SVM & 13.4   & 38.2 \\
		Our unsupervised +SVM & 40.7   & 60.9 \\
		Our supervised model & {\bf 52.9}  & {\bf 70.5} \\
		\hline 
		\textit{With ImageNet Pretrain}\\
		\hline
		5-layer Convnet~\citep{Zeiler14ECCV} &  {65}  & {74.2}\\
		5-layer Convnet~\citep{return} &-  &77.61\\
		Our supervised model   & {\bf 67.0}  & {\bf 77.9}  \\
		\hline
	\end{tabular}
	\label{Table:caltech256}
	\vspace{-1mm}
\end{table}

The state-of-the-art results on these two datasets are achieved by pretraining the deep network on a large dataset, ImageNet~\citep{decaf,HeSSPnet,Zeiler14ECCV}. We consider similar ImageNet pretraining in Sec.~\ref{Sec:ImageNet}.
We also observe from Table~\ref{Table:caltech256} that when there are fewer training images, our accuracy diminishes. This verifies that the model complexity needs to be selected based on the size of the data. This is also consistent with the results reported by~\citet{Zeiler14ECCV}, in which the classification performance is very poor without training the model on ImageNet.

\begin{figure}[tbp!]
	\centering
	\includegraphics[trim = 14.3cm  0 0 0.43cm, clip, width=0.445\textwidth] {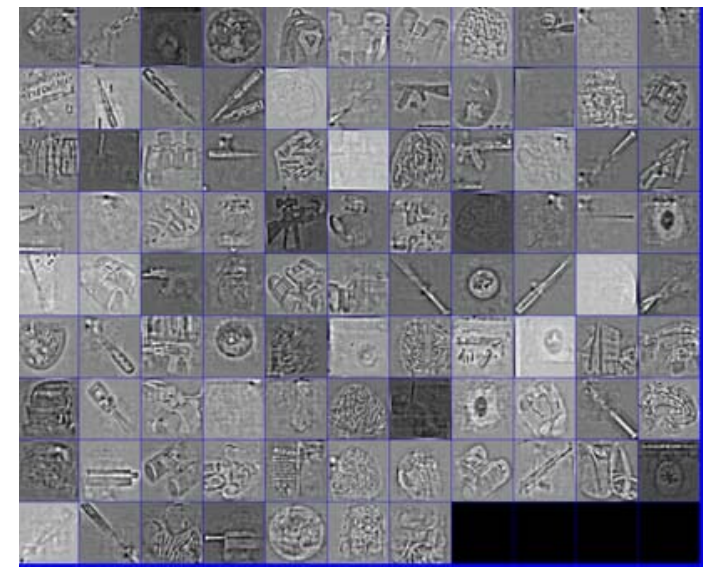}\\
	\includegraphics[trim = 14.3cm 0 0 0.45cm, clip,width=0.44\textwidth] {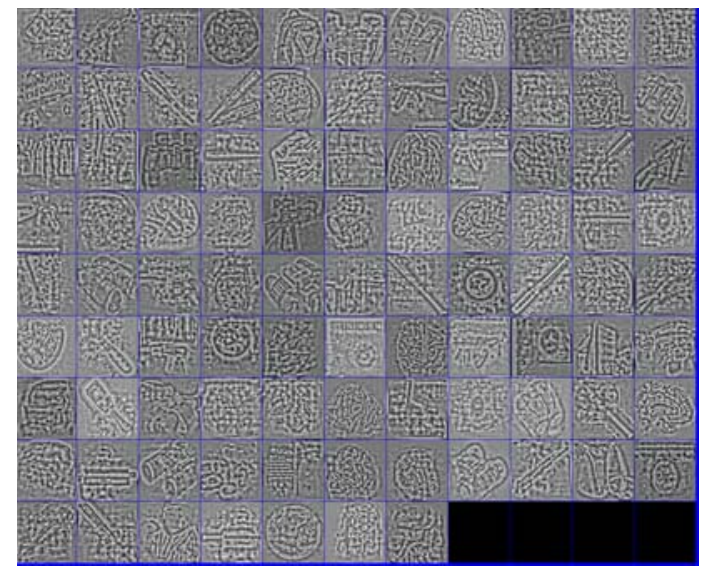}
	\vspace{-3mm}
	\caption{\small{Selected dictionary elements of our unsupervised model (top) and supervised model (bottom) trained from Caltech 256.}}
	\vspace{-5mm}
	\label{fig:super_unsper}
\end{figure}
\begin{figure}[htbp!]
	\centering
	\vspace{-1mm}
	\includegraphics[width=0.48\textwidth]{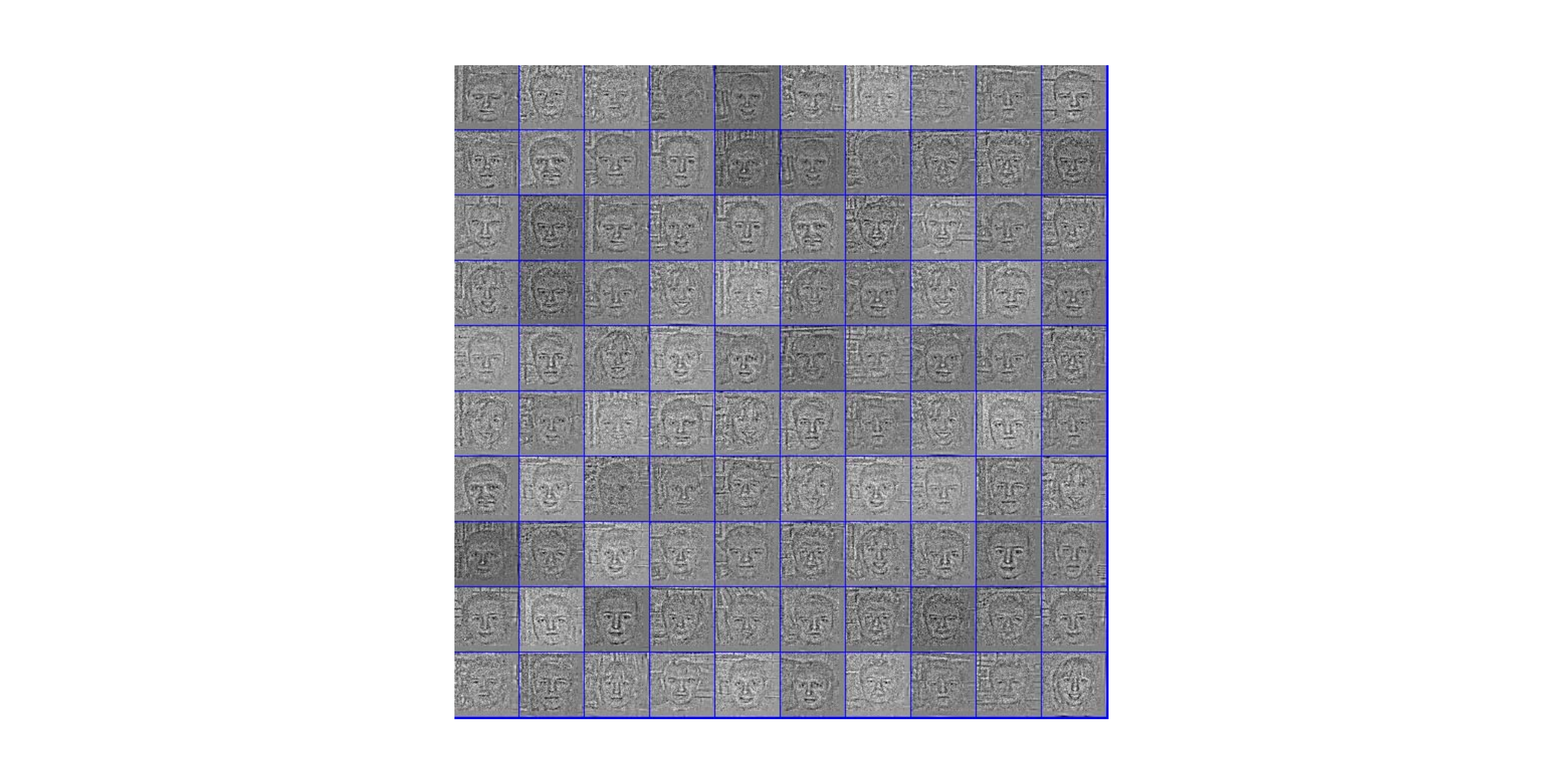}
	\vspace{-6mm}
	\caption{\small{Generated image from the dictionaries trained from the ``Faces\_easy" category using random dictionary weigths.}}
	\vspace{-3mm}
	\label{fig:face}
\end{figure}
\begin{figure}[htbp!]
	\centering
	\includegraphics[width=0.48\textwidth]{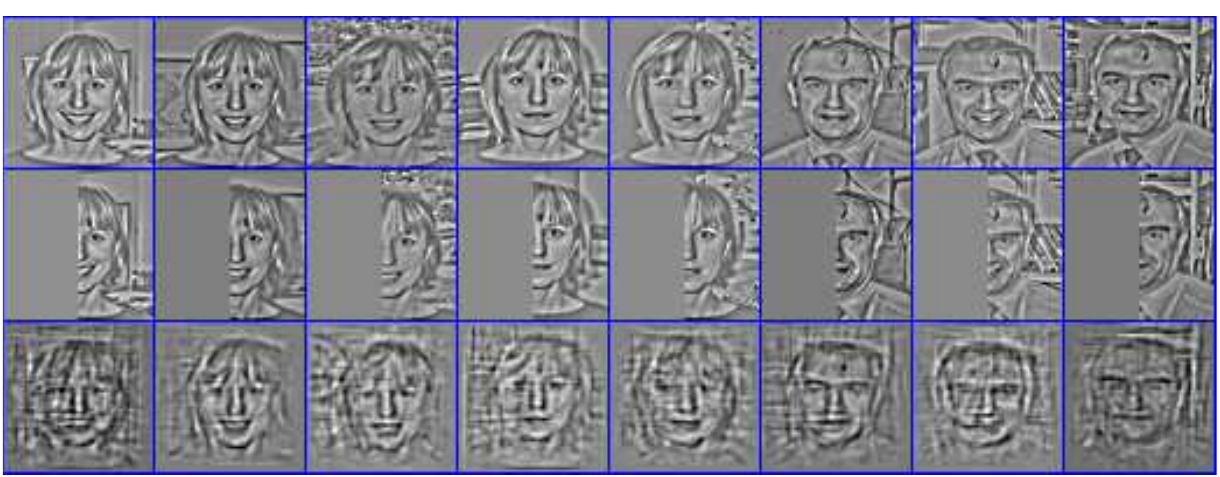}
	\vspace{-6mm}
	\caption{\small{Missing data interpolation. (Top) Original data. (Middle) Observed data. (Bottom) Reconstruction.}}
	\label{fig:face_missing}
	\vspace{-4mm}
\end{figure}
{Fig.~\ref{fig:super_unsper} shows selected dictionary elements learned from the unsupervised and the supervised model, to illustrate the differences. It is observed that the dictionaries without supervision tend to reconstruct the data while the dictionary elements with supervision tend to extract features that will distinguish different classes. For example, the dictionaries learned with supervision have double sides on the image edges.} Our model is generative, and as
an example we generate images using the dictionaries trained from the ``Faces easy" category, with random top-layer dictionary weights (see Fig.~\ref{fig:face}).
Similar to the MNIST example, we also show  in Fig.~\ref{fig:face_missing} the interpolation results of face data with half the image missing. Though the background is a little noisy, each face is recovered in great detail by the third (top) layer dictionaries. More results are provided in the SM.

\subsection{ImageNet 2012 \label{Sec:ImageNet}}
\vspace{-2mm}
We train our model on the 1000-category ImageNet 2012 dataset, which consists of 1.3M/50K/100K training/validation/test images. Our training process follows the procedure of previous work~\citep{howard2013some,HintonNIPS2012, Zeiler14ECCV}. The smaller image dimension is scaled to 256, and a $224\times 224$ crop is chosen at 1024 random locations within the image. The data are augmented by color alteration and horizontal flips \citep{howard2013some, HintonNIPS2012}. A five layer convolutional model is employed ($L=5$); the numbers (sizes) of dictionary elements for each layer are set to $96\,(5\times 5)$, $256\,(5\times 5)$, $512\,(3\times 3)$, $1024\,(3\times 3)$ and $\,512(3\times 3)$; the pooling ratios are $4\times 4\,$ (layer 1 to 2) and $2\times 2\,$ (others). The number of parameters in our model is around 30 million.  

We emphasize that our intention is not to directly compete with the best performance in the ImageNet challenge~\citep{googlenet,vgg}, which requires consideration of many additional aspects, but to provide a comparison on this dataset with a CNN with a similar network architecture (size). Table~\ref{Table:Error_Imagenet} summarizes our results compared with the ``ZF"-net developed in~\citet{Zeiler14ECCV} which has a similiar architecture with ours. 

The MAP estimator of our model, described in Sec.~\ref{Sec:train}, achieves a top-5 error rate of $16.1\%$ on the testing set, which is close to \citet{Zeiler14ECCV}. 
Model averaging used in Bayesian inference often improves performance, and is considered here.
Specifically, after running the MCEM algorithm, we have a (point) estimate of the global parameters. Using a mini-batch of data, one can leverage our analytic Gibbs updates to sample from the posterior (starting from the MAP estimate), and therefore obtain multiple samples for the global model parameters. We collect the approximate posterior samples every 1000 iterations, and retain 20 samples. 
Averaging the predictions of these 20 samples (model averaging) gives a top-5 error rate of $13.6\%$, which outperforms the combination of 6 ``ZF"-nets.
Limited additional training time (one day) is required for this model averaging.
\begin{table}[htbp!]
	\small
	\vspace{-2mm}
	\caption{ImageNet 2012 classification error rates ($\%$)}
	\vspace{-2mm}
	\centering
	\begin{tabular}{c|c|c}
		\hline
		Method & 	$\begin{array}{c}
		\text{top-1}\\
		\text{val}
		\end{array}$ & $\begin{array}{c}
		\text{top-5}\\
		\text{val}
		\end{array}$ \\
		\hline
		Our supervised model & 37.9& 16.1\\
		``ZF"-net~\citep{Zeiler14ECCV}&	37.5 & 16.0 \\	
		Our model averaging & {\bf 35.4} & {\bf 13.6} \\
		6 ``ZF"-net~\citep{Zeiler14ECCV}& 36 & 14.7 \\
		\hline       
	\end{tabular}\label{Table:Error_Imagenet}
	\vspace{-4mm}
\end{table}

To illustrate that our model can generalize to other datasets, we follow the setup in \citep{decaf, HeSSPnet,Zeiler14ECCV}, keeping five convolutional layers of our ImageNet-trained model fixed and train a new Bayesian SVM classifier on the top using the training images of Caltech 101 and Caltech 256, with each image resized to $256 \times 256$ (effectively, we are using ImageNet to pretrain the model, which is then refined for Caltech 101 and 256). The results are shown in Tables~\ref{Table:caltech101} and~\ref{Table:caltech256}. We obtain state-of-art results ($77.9\%$) on Caltech 256. For Caltech 101, our result ($93.15\%$) is competitive with the state-of-the-art result ($94.11\%$), which combines spatial pyramid matching and deep convolutional networks~\citep{HeSSPnet}. These results demonstrate that we can provide comparable results to the CNN in data generalization tasks, while also scaling well. 

\vspace{-2mm}
\section{Conclusions}
\vspace{-3mm}
A supervised deep convolutional dictionary-learning model has been proposed within a generative framework, integrating the Bayesian support vector machine and a new form of stochastic unpooling. Extensive image classification experiments demonstrate excellent classification performance on both small and large datasets. The top-down  form of the model constitutes a
new generative form of the deep deconvolutional network (DN) \citep{Zeiler10CVPR}, with unique learning and inference methods. 

\bibliographystyle{natbib}


%
%
%
%

\clearpage
\newpage 
\onecolumn
\section{More Results}
\subsection{Gnerated images with random weights}
\begin{figure}[!htbp]
	\centering
	\includegraphics[scale=0.6]{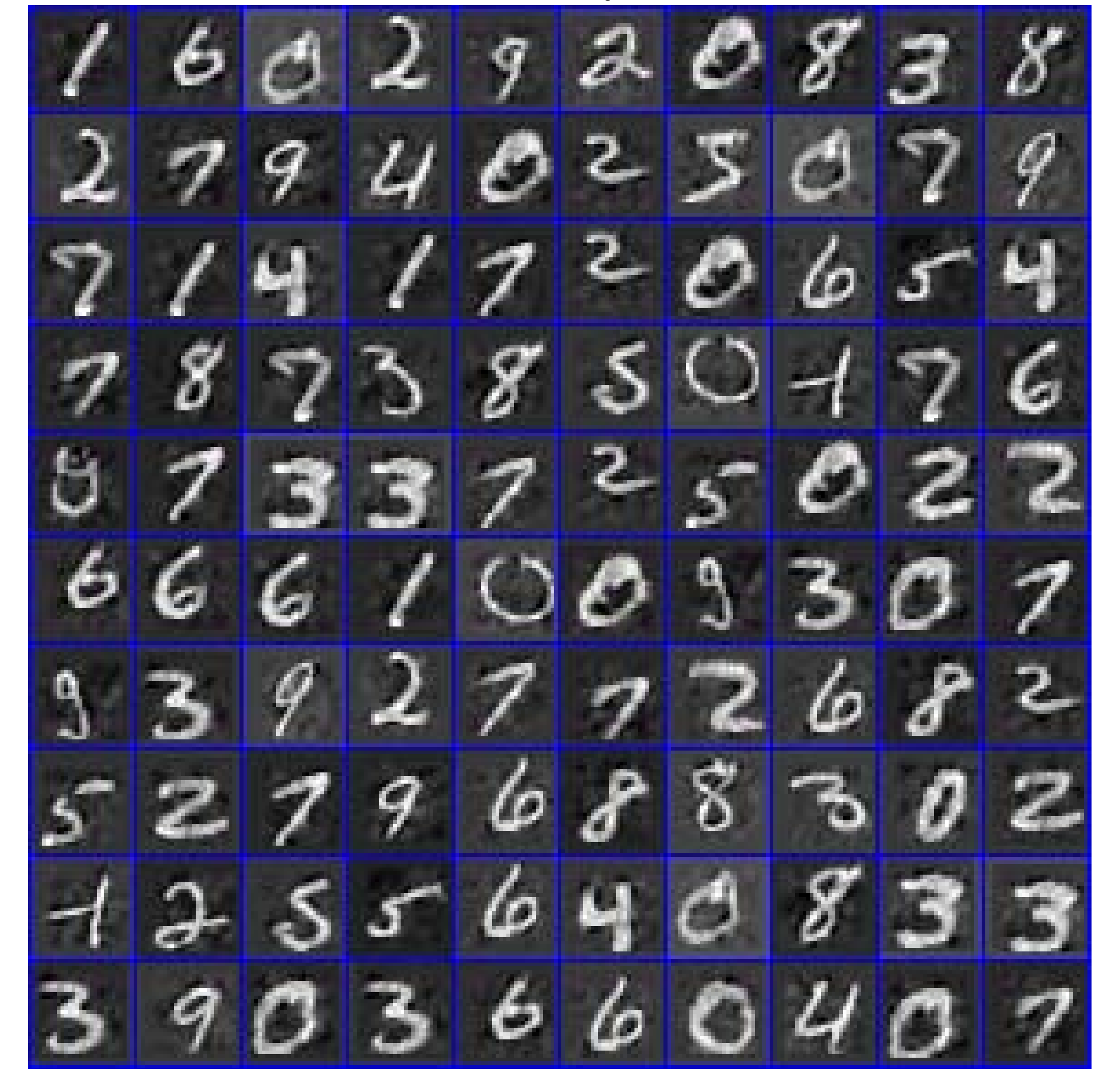}
	\caption{Generated images from the dictionaries trained from MNIST with random dictionary weights at the top of the two-layer model.}\label{fig:MNIST_Generate_all} 
\end{figure}
\clearpage
\newpage
\begin{figure}[!htbp]
	\centering
	\includegraphics[width=\textwidth]{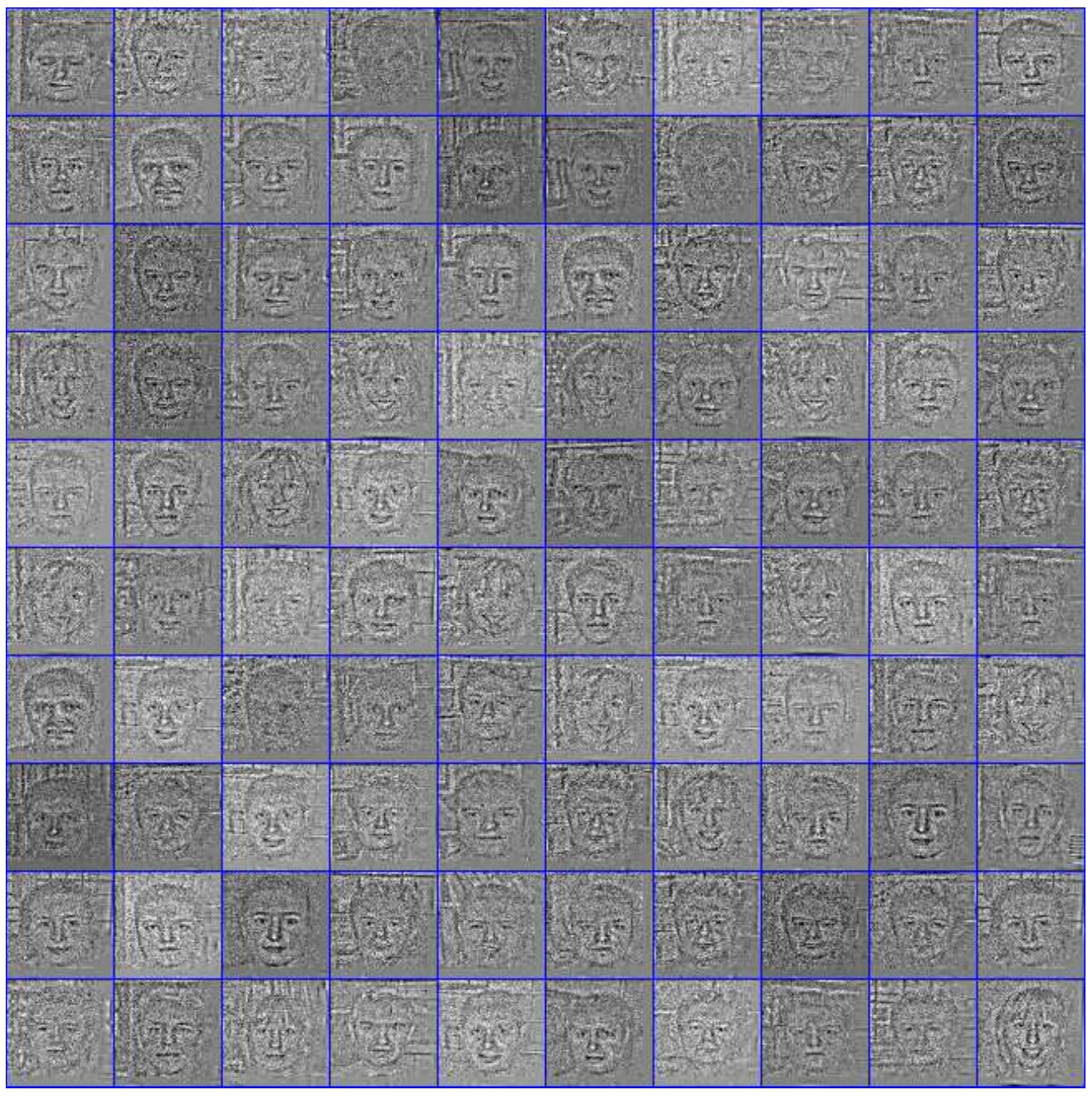}
	\caption{Generated images from the dictionaries trained from ``Faces\_easy" category of Caltech 256 with random dictionary weights at the top of the three-layer model.}\label{fig:Face_Generate_all}
\end{figure}
\clearpage
\newpage
\begin{figure}[!htbp]
	\centering
	\includegraphics[width=\textwidth]{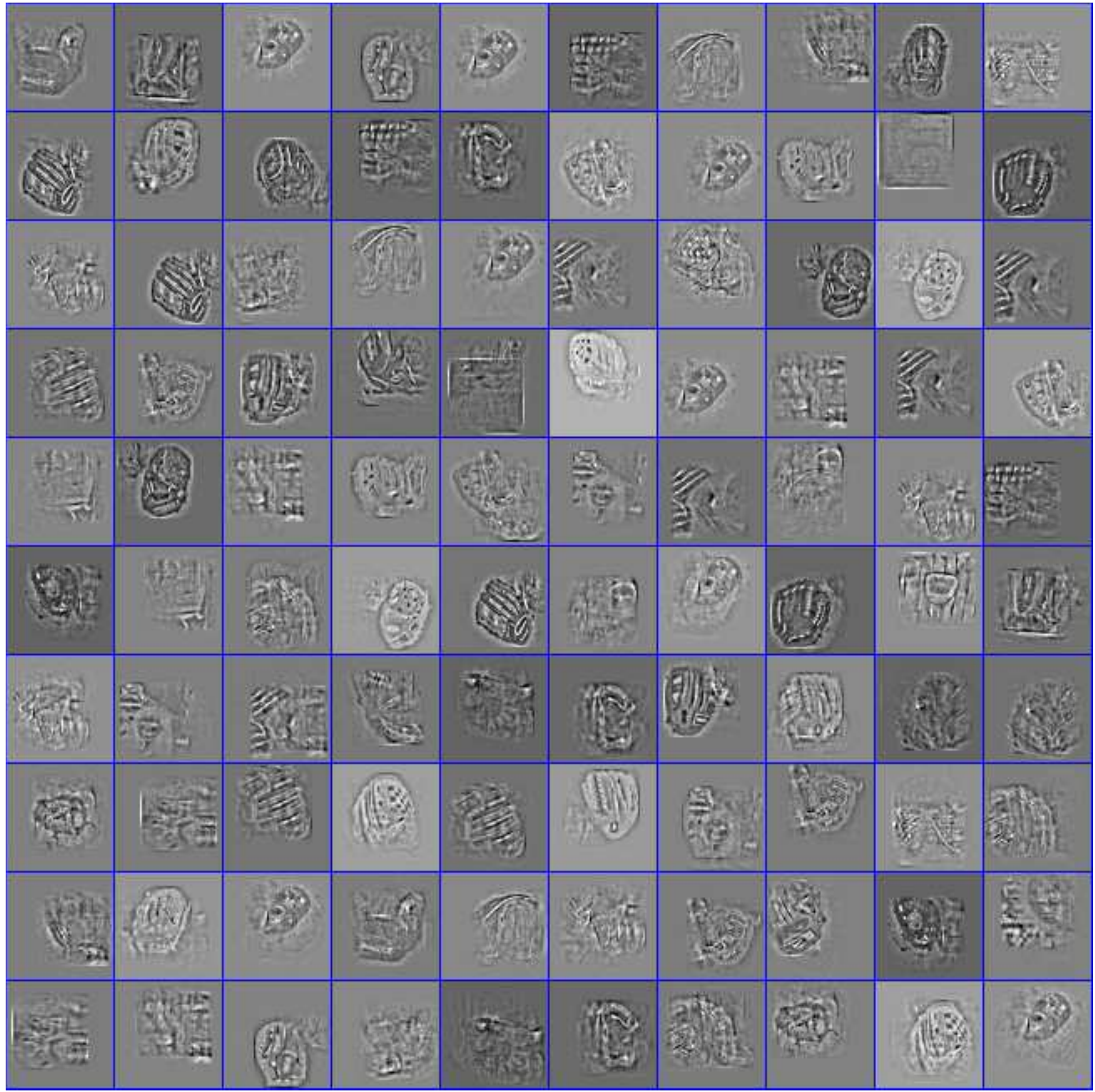}
	\caption{Generated images from the dictionaries trained from `baseball-glove" category of Caltech 256 with random dictionary weights at the top of the three-layer model.}\label{fig:backpack_Generate_all}
\end{figure}
\clearpage
\subsection{Missing data interpolation}
\begin{figure}[!htbp]
	\centering
	\includegraphics[scale=0.9]{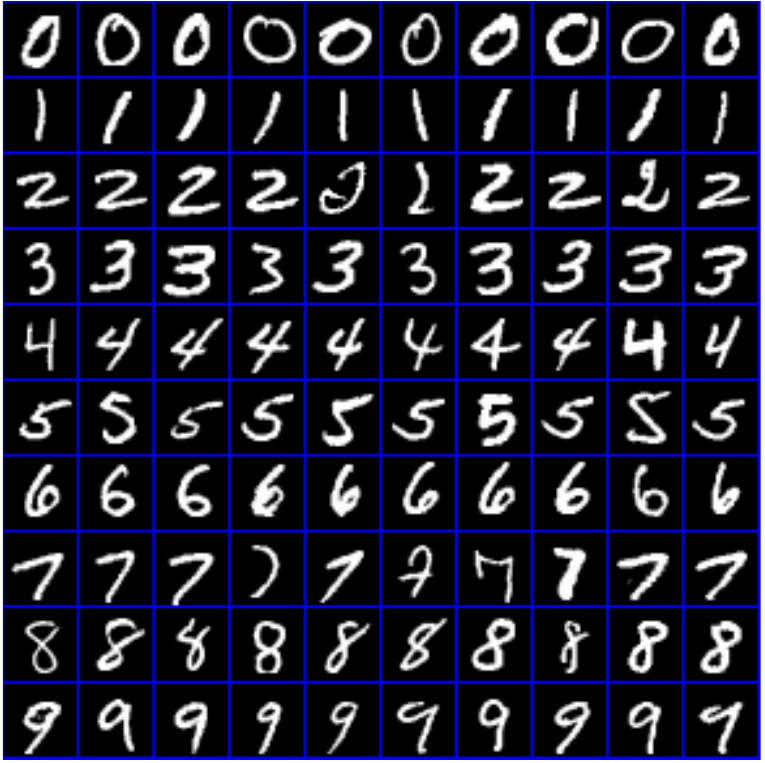}\hspace{2cm}
	\includegraphics[scale=0.75]{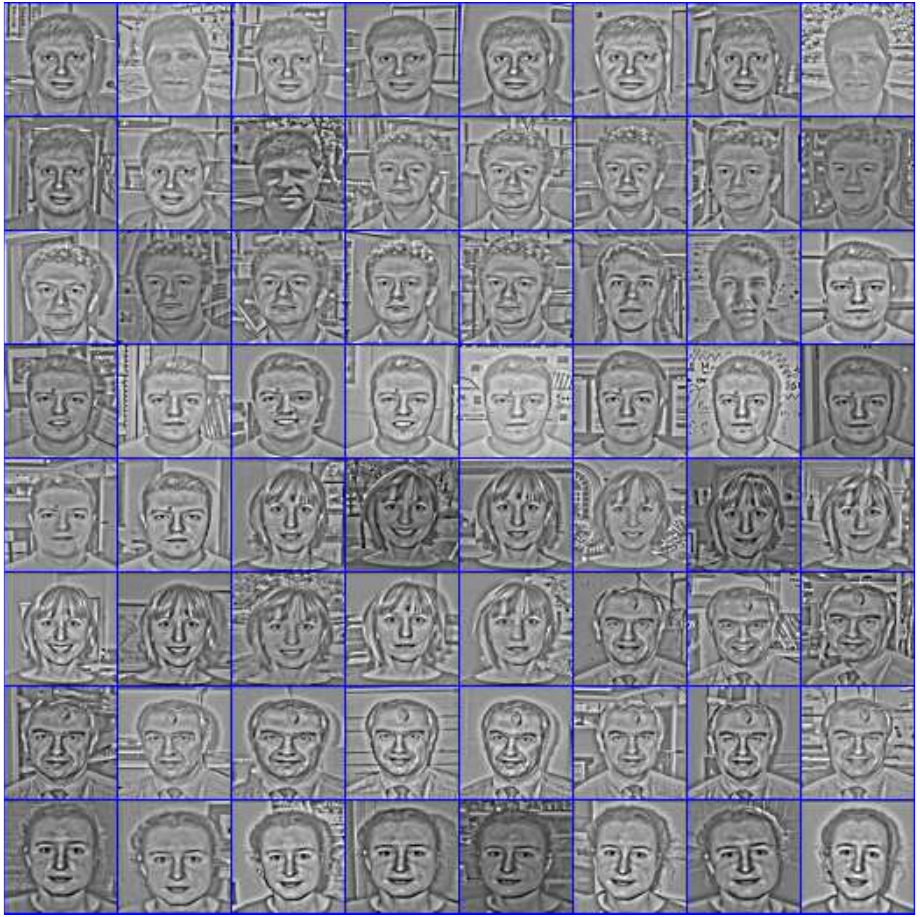}\\
	\includegraphics[scale=0.9]{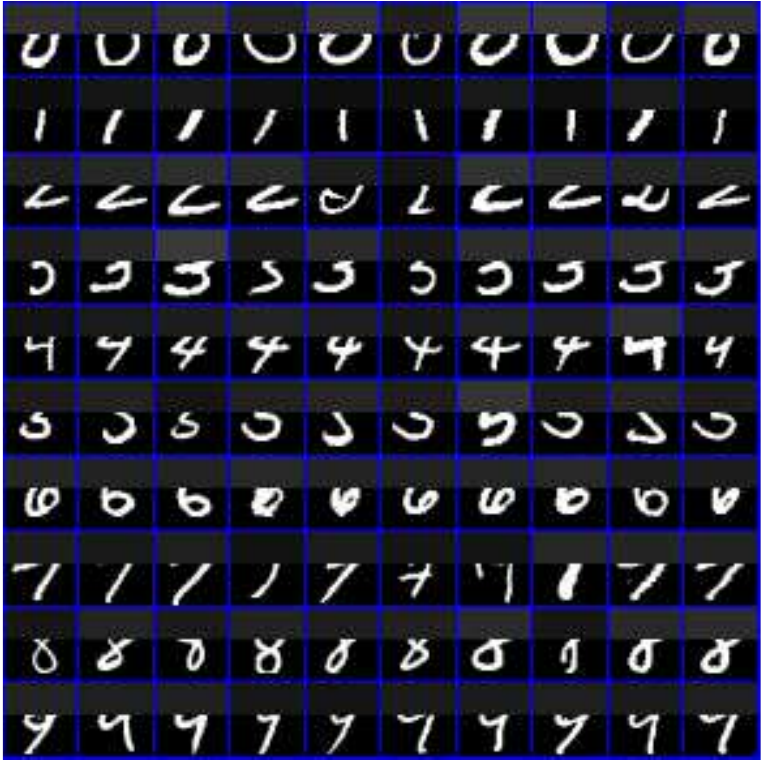}\hspace{2cm}
	\includegraphics[scale=0.75]{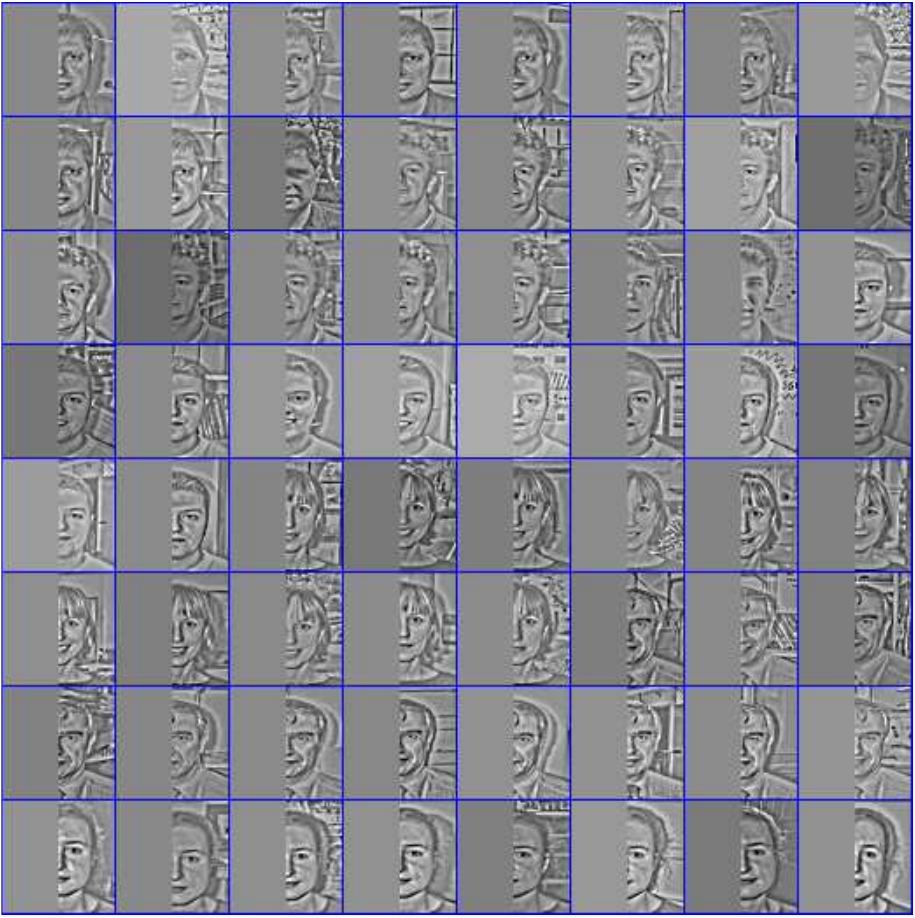}\\
	\hspace{0.1cm}\includegraphics[scale=0.9]{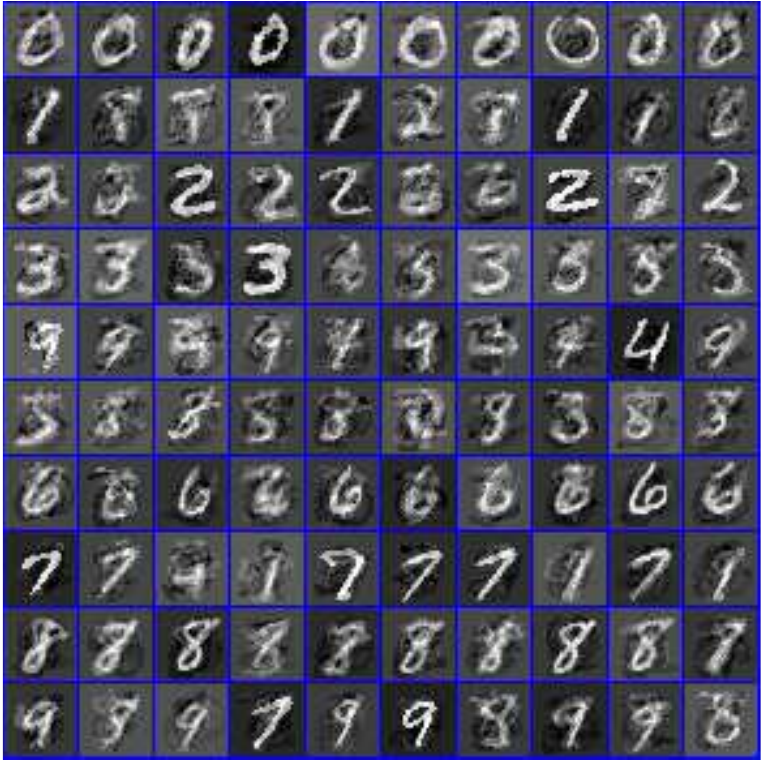}\hspace{1.95cm}
	\includegraphics[scale=0.75]{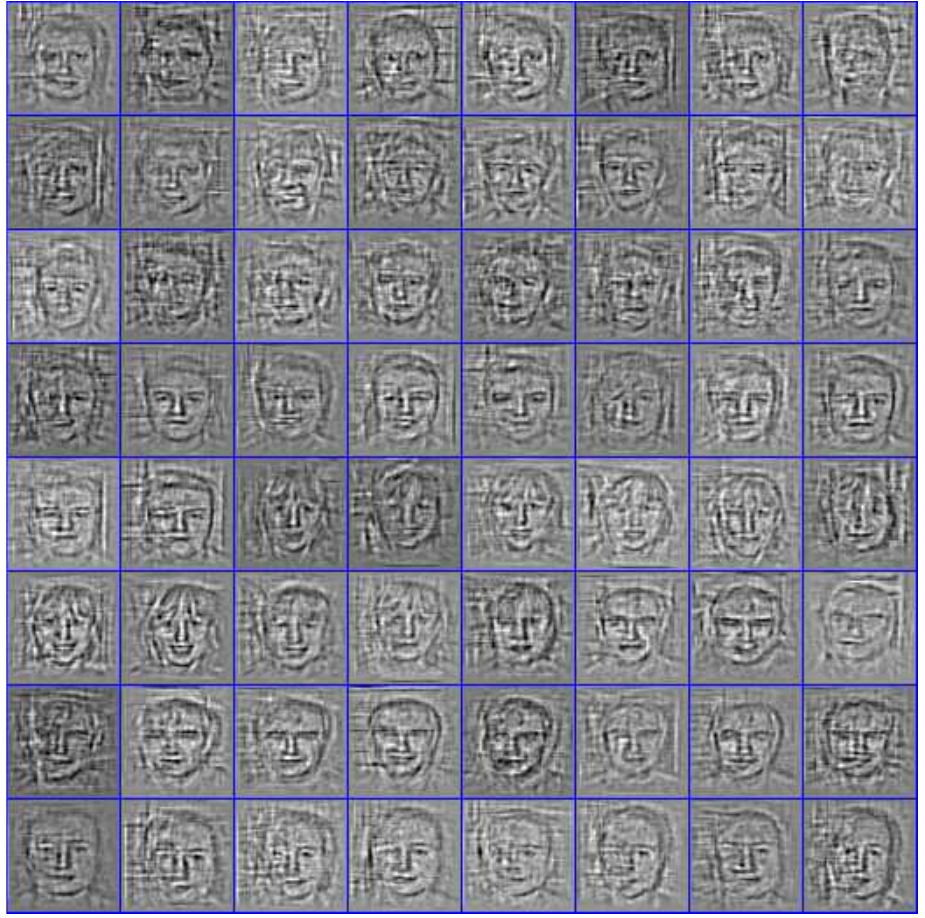}
	\caption{Missing data interpolation of digits (left column) and Face easy (right column). For each column: (Top) Original data. (Middle)
		Observed data. (Bottom) Reconstruction.}\label{fig:MNIST_missing_all}
\end{figure}	
\clearpage
\newpage
\section{MCEM algorithm}
Algorithms \ref{algo:onlineMCEM} and \ref{algo:test} detail the training and testing process. The steps are explained in the next two sections.

\begin{center}
	\begin{algorithm}[h!]
		\caption{Stochastic MCEM Algorithm}
		\begin{algorithmic}
			\REQUIRE Input data $\{{\Xmat}^{(n)},\ell_n\}_{n=1}^N$.
			\FOR{$t=1$ \TO $\infty$}
			\STATE Get mini-batch (${\Ymat}^{(n)};n\in \mathcal{I}_t$) randomly.
			\FOR{$s=1$ \TO $N_s$}
			\STATE Sample $\{\gammav_e^{(n,k_0)}\}_{k_0=1}^{K_0}$ from the distribution in (\ref{Eq:SampleE}); \\sample  $\{{\gamma_s}^{(n,k_L)}\}_{k_L=1}^{K_L}$ from the distribution in (\ref{Eq:SampleR});\\
			sample $\{\{\Zmat^{(n,k_l,l)}\}_{k_l=1}^{K_l}\}_{l=1}^{L}$ from the distribution in (\ref{Eq:SampleZ});	\\sample $\{{\Smat}^{(n,k_L,L)}\}_{k_L=1}^{K_L}$ from the distribution in (\ref{Eq:SampleS}).
			\ENDFOR
			\STATE Compute $\bar{Q}(\Psimat|\Psimat^{(t)})$ according to (\ref{Eq:ApproQ})
			\FOR{$l=1$ \TO $L$}
			\STATE Update $\{\deltav^{(n,k_{l-1},l,t)}\}_{k_{l-1}=1}^{K_{l-1}}$ according to (\ref{Eq:UpdateDelta}).
			\FOR{$k_{l-1}=1$ \TO $K_{L-1}$}
			\FOR{$k_{l}=1$ \TO $K_{L}$}
			\STATE Update ${\Dmat}^{(k_{l-1},k_l,l,t)}$ according to (\ref{Eq:UpdateD}).
			\ENDFOR 
			\STATE 	Update 
			$\bar{\Xmat}^{(n,k_{l-1},l,t)}:=\sum_{k_l=1}^{K_l} {\Dmat}^{(k_{l-1},k_l,l,t)} * \bar {\Smat}^{(n,k_l,l,t)}$.
			\STATE Update $\bar{\Smat}^{(n,k_{l-1},l-1,t)}=f(\bar{\Xmat}^{(n,k_{l-1},l,t)}, \bar{\Zmat}^{(n,k_{l-1},l-1,t)} )$.
			\ENDFOR 
			\ENDFOR  
			\FOR{$\ell=1$ \TO $C$} 
			\STATE Sample $\lambda_n^{(\ell)}$ from the distibution in (\ref{Eq:Samplelambda}) and compute the sample average $\bar{\lambda}_n^{(\ell,t)}$.
			\STATE Update $\betav^{(\ell,t)}$ according to (\ref{Eq:UpdateBeta}).
			\ENDFOR  	
			
			\ENDFOR
			\RETURN A point estimator of $\Dmat$ and $\betav$.
		\end{algorithmic}
		\label{algo:onlineMCEM}
	\end{algorithm}
\end{center}
\vspace{-5mm}
\begin{center}
	\begin{algorithm}[h!]
		\caption{Testing}
		\begin{algorithmic}
			\REQUIRE Input test images ${\Xmat}^{(*)}$, learned dictionaires $\{\{\Dmat^{(k_l,l)}\}_{k_l=1}^{K_L}\}_{l=1}^{L}$
			\FOR{$t=1$ \TO $T$}
			\FOR{$s=1$ \TO $N_s$}
			 \STATE Sample $\{\gammav_e^{(n,k_0)}\}_{k_0=1}^{K_0}$ from the distribution in (\ref{Eq:SampleE}); \\sample  $\{{\gamma_s}^{(n,k_L)}\}_{k_L=1}^{K_L}$ from the distribution in (\ref{Eq:SampleR});\\
			 sample $\{\{\Zmat^{(n,k_l,l)}\}_{k_l=1}^{K_l}\}_{l=1}^{L-1}$ from the distribution in (\ref{Eq:SampleZ});\\
			\ENDFOR
			\STATE Compute $\bar{Q}_{test}(\Psimat_{test}|\Psimat_{test}^{(t)})$ according to (\ref{Eq:ApproTestQ})
			\FOR{$l=1$ \TO $L$}
			\STATE Update $\{\deltav^{(*,k_{l-1},l,t)}\}_{k_{l-1}=1}^{K_{l-1}}$ according to (\ref{Eq:UpdateDelta}).
			\ENDFOR  
			\FOR{$k_L=1$ \TO $K_L$}
			\STATE Update ${\Zmat}^{(*,k_L,L)}$ according to (\ref{Eq:UpdateZ}).
			\STATE Update ${\Wmat}^{(*,k_L,L)}$ according to (\ref{Eq:UpdateW}).	
			\ENDFOR	
			\ENDFOR  
			\STATE Compute $\{{\Smat}^{(*,k_L,L)}\}_{k_L=1}^{K_L}$ and get its vector verstion $\sv_{*}$.
			\STATE Predict label $\ell^*$=$\arg\max_{\ell}\betav_{\ell}^\top\sv_{*}$.	
			\RETURN the predicted label  $\ell^*$ and the decision value $\betav_{\ell}^\top\sv_{*}$.
		\end{algorithmic}
		\label{algo:test}
	\end{algorithm}
\end{center}
\section{Gibbs Sampling}
\subsection{Notations}
In the remainder of this discussion, we use the following definitions.
\begin{enumerate}[(1)]	
	\item
	{\bf{The ceiling function}}: 
	
	$ceil(x) =\lceil x\rceil$ is the smallest integer that is not less than $x$.
	\item
	{\bf {The summation and the quadratic summation of all elements in a matrix}}:
	
	if $\Xmat\in \mathbb{R}^{N_x\times N_y}$,
	\begin{align}
	\mathsf{sum}(\Xmat) =\sum_{i=1}^{N_x}\sum_{j=1}^{N_y}X_{ij}, \qquad\qquad\|\Xmat\|_2^2=\sum_{i=1}^{N_x}\sum_{j=1}^{N_y}X_{ij}^2.
	\end{align}
	\item
	{\bf{The unpooling function}}:
	
	Assume ${\Smat}\in {\mathbb R}^{N_x\times N_y}$ and ${\Xmat}\in {\mathbb R}^{N_x/p_x\times N_y/p_y}$. Here $p_x, p_y\in N$ are the pooling ratio and the pooling map is ${\Zmat}\in \{0,1\}^{N_x\times N_y}$. Let $i^{\prime}\in\{1,...,\lceil N_x/p_x \rceil\}$, $j^{\prime}\in\{1,...,\lceil N_y/p_y\rceil\}$, $i\in\{1,...,N_x\}$, $j\in\{1,...,N_y\}$, then
	$f:{\mathbb R}^{Nx/p_x\times Ny/p_y}\times \{0,1\}^{N_x\times N_y}\to {\mathbb R}^{N_x\times N_y}$.
	
	If ${\Smat}=f({\Xmat},{\Zmat})$
	\begin{align}
	S_{i,j}=X_{\left\lceil i/p_x\right\rceil, \left\lceil j/p_y\right\rceil}Z_{i,j}.
	\end{align}
	Thus, the unpooling process (equation(6) in the main paper) can be formed as:
	\begin{align}
	\Smat^{(n,k_l,l)}=\mathsf{unpool}(\Xmat^{(n,k_l,l+1)})=f(\Xmat^{(n,k_l,l+1)},\Zmat^{(n,k_l,l)}).
	\end{align}
	\item
	{\bf{The 2D correlation operation}}:
	
	Assume ${\Bmat} \in {\mathbb R}^{N_{Bx} \times N_{By}}$ and ${\Cmat} \in {\mathbb R}^{N_{Cx} \times N_{Cy}}$. If ${\Amat}={\Bmat}\circledast {\Cmat}$, then  ${\Amat} \in {\mathbb R}^{(N_{Bx}-N_{Cx}+1) \times (N_{By}-N_{Cy}+1)}$ with element $(i,j)$ given by
	\begin{align}
	A_{i,j} = \sum_{p=1}^{N_{Cx}} \sum_{q=1}^{N_{Cy}} B_{p+i-1,q+j-1} C_{p,q}.
	\end{align}
	\item
	{\bf{The ``error term" in each layer}}:
	\begin{align}
	    \delta_{i,j}^{(n,k_{l-1},l)}=\frac{\partial}{\partial {\Xmat}^{(n,k_{l-1},l)}_{i,j}}\left\{\frac{\gamma^{(n)}_e}{2}\sum_{k_0=1}^{K_0}\|\Emat^{(n,k_0)}\|_2^2\right\}.
	\end{align}
	\item 
	{\bf{The ``generative" function}}:
	
	This  ``generative" function measures how much the $k^{\rm th}$ band of  $l^{\rm th}$ layer feature is ``responsible" for the of input image ${\Xmat}^{(n)}$ in the current model:
	\begin{equation}
	g({\Xmat},n,k,l) =\begin{cases}
	{\Dmat}^{(k,1)} * f({\Xmat}, {\Zmat}^{(n,k,1)}) &\text{if}\ l=2, \\
	\sum_{m=1}^{K_{l-1}}g\left({\Dmat}^{(m,k,l-1)} * f({\Xmat}, {\Zmat}^{(n,k,l-1)}),n,m,l-1\right)  & \text{if}\ l>2.
	\end{cases}
	\end{equation}
	It can be considered as if $k^{\rm th}$ band of  $l^{\rm th}$ layer feature changes $\Xmat$ (\ie \, $\Xmat^{(n,k,l)}\rightarrow \Xmat^{(n,k,l)}+\Xmat$), the corresponding data layer representation will change $g({\Xmat},n,k,l)$ (\ie \, $\Xmat^{(n)}\rightarrow \Xmat^{(n)}+g({\Xmat},n,k,l)$).  Thus, for $l=2,\dots,L$, we have
	\begin{align}
	\Xmat^{(n)}=\sum_{k=1}^{K_l}g(\Xmat,n,k,l)+\Emat^{(n)}.
	\end{align}
	
	Note that $g()$ is a {\em linear} function for $\Xmat$, which means:
	\begin{align}
	g({\mu_1\Xmat_1+\mu_2\Xmat_2},n,k,l)=\mu_1g({\Xmat_1},n,k,l)+\mu_2g({\Xmat_2},n,k,l).
	\end{align}
\end{enumerate}	
For convenience, we also use the following notations:
\begin{itemize}
\item We use ${\Zmat}^{(n,k_l,l)}$ to represent $\{\zv^{(n,k_l,l)}_{i,j};\forall i,j\}$, where the vector version of the $(i,j)^{\rm th}$ block of ${\Zmat}^{(n,k_l,l)}$ is equal to  $\zv^{(n,k_l,l)}_{i,j}$. 
\item
$\bf{0}$ denotes the all 0 vector or matrix. $\bf{1}$ denotes the all one vector or matrix. $\ev_m$ denotes a ``one-hot" vector with the $m^{\rm th}$ element equal to 1.
\end{itemize}

\subsection{Full Conditional Posterior Distribution}\label{sec:posterior}

Assume the spatial dimension: ${\Xmat}^{(n,l)}\in \mathbb{R}^{N_{x}^l \times N_{y}^l\times K_{l-1}}$, ${\Dmat}^{(k_l,l)}\in \mathbb{R}^{N_{dx}^l \times N_{dy}^l\times K_{l-1}}$, ${\Smat}^{(n,k_l,l)}\in \mathbb{R}^{N_{Sx}^l \times N_{Sy}^l}$ and ${\Zmat}^{(n,k_l,l)}\in \{0,1\}^{N_{Sx}^l \times N_{Sy}^l}$.  
For  $\ l=0,\dots,L$, we have $k_l=1,\dots,K_l\ $. The (un)pooling ratio from $l-$th layer to $(l+1)-$layer is $p_x^l\times p_y^l$ (where $l=1,\dots,L-1$). We have:
\begin{align}
N_{x}^l&=N_{dx}^l+N_{Sx}^l-1,& N_{Sx}^l& =p_x^l\times 	N_{x}^{(l+1)},\\
N_{y}^l&=N_{dy}^l+N_{Sy}^l-1, & N_{Sy}^l& =p_y^l\times 	N_{y}^{(l+1)}.
\end{align} 
Recall that, for $l=2,\dots,L$:
\begin{align}
\Xmat^{(n,k_{l-1},l)}=\sum_{k_l}^{K_l}\Dmat^{(k_{l-1},k_l,l)}*\Smat^{(n,k_{l},l)}.
\end{align}
Without loss of generality, we omit the superscript $(n,k_{l-1},l)$ below. Each element of $\Xmat$ can be represent as:
\begin{eqnarray}
X_{i,j}&=&\sum_{p=1}^{N_{dx}}\sum_{q=1}^{N_{dy}}D_{p,q}S_{(i+N_{dx}-p,j+N_{dy}-q)}\nonumber \\
&=& D_{p,q}S_{(i+N_{dx}-p,j+N_{dy}-q)}+X^{-(p,q)}_{i,j}
\end{eqnarray}
where $X^{-(p,q)}_{i,j}$ is a term which is independent of $D_{p,q}$ but related by the index $(i,j,p,q)$; so is $S_{(i+N_{dx}-p,j+N_{dy}-q)}$. Following this, for every elements in $\Dmat$, we can represent $\Xmat$ as:
\begin{align}
	\Xmat=\Xmat_{-(p,q)}+D_{p,q}\Smat_{-(p,q)}
\end{align}
where matrices $\Xmat_{-(p,q)}$ and $\Smat_{-(p,q)}$ are independent of $D_{p,q}$ but related by the index $(p,q)$ (and the superscript $(n,k_{l-1},l)$).
Therefore:
\begin{align}
{\Emat}^{(n)}&=\Xmat^{(n)}-\sum_{k=1}^{K_l}g(\Xmat,n,k,l)\\
&=\Xmat^{(n)}-\sum_{k=1,\neq k_{l-1}}^{K_l}g(\Xmat,n,k,l)-g(\Xmat,n,k_{l-1},l)\\
&=\Xmat^{(n)}-\sum_{k=1,\neq k_{l-1}}^{K_l}g(\Xmat,n,k,l)-g\left(\Xmat_{-(p,q)}+D_{p,q}\Smat_{-(p,q)},n,k_{l-1},l\right)\\
&=\Xmat^{(n)}-\sum_{k=1,\neq k_{l-1}}^{K_l}g(\Xmat,n,k,l)-g\left(\Xmat_{-(p,q)},n,k_{l-1},l\right)+g\left(\Smat_{-(p,q)},n,k_{l-1},l\right)D_{p,q}\\
&=\Cmat_{p,q}-D_{p,q}\Fmat_{(p,q)}
\end{align}
If we add the superscripts back, we have:
\begin{align}
{\Emat}^{(n)}={\Cmat}^{(n,k_{l},l)}_{p,q}+{D}^{(n,k_{l},l)}_{p,q}{\Fmat}^{(n,k_{l},l)}_{p,q},
\end{align}
where matrices ${\Cmat}^{(n,k_{l},l)}_{p,q}$ and ${\Fmat}^{(n,k_{l},l)}_{i,j}$ are independent of ${D}^{(n,k_{l},l)}_{p,q}$ but related by the index $(n,k_{l},l,p,q)$.

Similarly, for every elements in $\zv$, we have
\begin{align}
{\Emat}^{(n)}={\Amat}^{(n,k_{l},l)}_{i,j,m}+{z}^{(n,k_{l},l)}_{i,j,m}{\Bmat}^{(n,k_{l},l)}_{i,j,m}.
\end{align}

\begin{enumerate}	
	\item The conditional posterior of $\Dmat_{i,j}^{(k_{l-1},k_l,l)}$:
	\begin{eqnarray}\label{Eq:SampleD}
		D_{i,j}^{(k_{l-1},k_l,l)}|- &\sim&{\mathcal N} (\mu_{i,j}^{(k_{l-1},k_l,l)}, \sigma_{i,j}^{(k_{l-1},k_l,l)}),
		\end{eqnarray}
		where 
		\begin{eqnarray}
		\sigma_{i,j}^{(k_{l-1},k_l,l)}&=&\left(\frac{\gamma_e^{n}}{2}\|{\Fmat}^{(n,k_{l},l)}_{i,j}\|_2^2+1\right)^{-1},\\
		\mu_{i,j}^{(k_{l-1},k_l,l)}&=&\sigma_{i,j}^{(k_{l-1},k_l,l)}\mathsf{sum}({\Cmat}^{(n,k_{l},l)}_{i,j}\circ{\Fmat}^{(n,k_{l},l)}_{i,j}).
		\end{eqnarray}
	\item The conditional posterior of $\zv_{i,j}^{(n,k_l,l)}$:
	\begin{align}\label{Eq:SampleZ}
		\zv_{i,j}|-&~\sim~~ \hat{\thetav}_0[\zv_{i,j}={\bf 0}]+\sum_{m=1}^{p_x^lp_y^l}\hat{\thetav}_m[\zv_{i,j}={\ev_m}],
	\end{align}
	where 
	\begin{align}
	\hat{\theta}_{m}&=\frac{\theta_{i,j}^{(m)}\eta_{i,j}^{(m)}}{\theta_{i,j}^{(0)}+\sum_{\hat{m}=1}^{p_xp_y} \theta_{i,j}^{(\hat m)} \eta_{i,j}^{(\hat{m})}},\\
	\hat{\theta}_{0}&=\frac{\theta_{i,j}^{(0)}}{\theta_{i,j}^{(0)}+\sum_{\hat{m}=1}^{p_xp_y} \theta_{i,j}^{(\hat m)} \eta_{i,j}^{(\hat{m})}},\\
	\eta_{i,j}^{(m)}&=\exp\left\{-\frac{{\gamma_e}}{2} \left(\|\Amat_{i,j}^{(m)} - \Bmat_{i,j}^{(m)}\|_2^2-\|\Amat_{i,j}^{(m)}\|_2^2\right)\right\}.\label{Eq:eta}
	\end{align}
	For notational simplicity, we omit the superscript $(n,k_l,l)$. We can see that when $\eta_{i,j}^{(m)}$ is large, $\hat{\theta}_{m}$ is large, causing the $m^\text{th}$ pixel to be activated as the unpooling location. When all of the $\eta_{i,j}^{(m)}$ are small the model will prefer not unpooling -- none of the positions $m$ make the model fit the data (\ie, $\Bmat_{i,j}^{(m)}$ is not close to $\Amat_{i,j}^{(m)}$ for all $m$); this is mentioned in the main paper.
	\item
	The conditional posterior of $\thetav^{(n,k_{l},l)}$
	\begin{eqnarray}
	\thetav^{(n,k_{l},l)} |- &\sim& \text{Dir} ({\alpha}^{(n,k_{l},l)}),
	\end{eqnarray}
	where
	\begin{align}
	\alpha_m^{(n,k_{l},l)} &= \frac{1}{p_x^lp_y^l+1} + \sum_{i} \sum_{j} Z_{i,j,m}^{(n,k_{l},l)} \qquad \text{for }~ m=1,...,p_x^lp_y^l,\\
	\alpha_{0}^{(n,k_{l},l)} &= \frac{1}{p_x^lp_y^l+1} + \sum_{i} \sum_{j} \left(1-\sum_m Z_{i,j,m}^{(n,k_{l},l)}\right).
	\end{align}
	\item
	The conditional posterior of $S^{(n,k_{L},L)}_{i,j}$:
	\begin{align}\label{Eq:SampleS}
	S^{(n,k_{L},L)}_{i,j}|- \sim(1-{Z}^{(n,k_L,L)}_{i,j})\delta_0 +{Z}^{(n,k_L,L)}_{i,j}{\mathcal N} (\Xi^{(n,k_{L},L)}_{i,j}, \Delta^{(n,k_{L},L)}_{i,j}),
	\end{align}
	where 
	\begin{align}
			\Delta^{(n,k_{L},L)}_{i,j}&= \left(\gamma_e^{(n)} \|\Fmat^{(n,k_{L},L)}_{i,j}Z^{(n,k_{L},L)}_{i,j} \|_2^2+\sum_{\ell}\frac{\gamma }{\lambda_n^{(\ell)}}y_n^{(\ell)}(Z^{(n,k_{L},L)}_{i,j}\hat{\beta}^{(k_L,\ell)}_{i,j})^2 + \gamma_s^{(n,k_{L})}\right)^{-1},\nonumber\\
			\Xi^{(n,k_{L},L)}_{i,j}&= \Delta^{(n,k_{L},L)}_{i,j}Z^{(n,k_{L},L)}_{i,j}\left(\mathsf{sum}(\Fmat^{(n,k_{L},L)}_{i,j}\circ \Cmat^{(n,k_{L},L)}_{i,j}) + \sum_{\ell}y_n^{(\ell)}\hat{\beta}^{(k_L,\ell)}_{i,j}(1+\lambda_n^{(\ell)})	\right).
	\end{align}
		Here we reshape the long vector $\betav_{\ell}\in \mathbb{R}^{N_{sx}^lN_{sy}^LK_L\times 1}$ into a matrix $\hat{\betav}_{\ell}\in \mathbb{R}^{N_{sx}^l\times N_{sy}^L\times K_L}$ which has the same size of ${\Smat}^{(n,L)}$.
		\item 
		The conditional posterior of $\gamma_s^{(n,k_L)}$:
		\begin{align}\label{Eq:SampleR}
		\gamma_s^{(n,k_L)}|- \sim \text{Gamma}\left(a_s+\frac{N_{Sx}^L\times N_{Sy}^L}{2},  b_s+\frac{1}{2}\|{\Smat}^{(n,k_{L},L)}\|_2^2\right).
		\end{align}
		\item
		The conditional posterior of $\gamma_e^{(n)}$:
		\begin{align}\label{Eq:SampleE}
			\gamma_e^{(n)}|- \sim \text{Gamma}\left(a_0+\frac{N_x\times N_y\times K_0}{2},  b_0+\frac{1}{2}\sum_{k_0=1}^{K_0} \|\Emat^{(n,k_0)}\|_2^2\right).
		\end{align}
		\item
		The conditional posterior of $\betav_\ell$:
		
		Reshape the long vector $\betav_{\ell}\in \mathbb{R}^{N_{sx}^lN_{sy}^LK_L\times 1}$ into a matrix $\hat{\betav}_{\ell}\in \mathbb{R}^{N_{sx}^l\times N_{sy}^L\times K_L}$ which has the same size as ${\Smat}^{(n,L)}$. We have:
		\begin{eqnarray}
		\displaystyle\hat{\beta}^{(k_L,\ell)}_{i,j}|-&\sim& {\cal N}(\mu^{(k_L,\ell)}_{i,j},\sigma^{(k_L,\ell)}_{i,j}),\\
		\sigma^{(k_L,\ell)}_{i,j}&=&\left(\sum_n\frac{\gamma }{\lambda_n^{(\ell)}}y_n^{(\ell)}(S_{i,j}^{(n,k_L,L)})^2+\frac{1}{\omega^{(k_L,\ell)}_{i,j}}\right)^{-1},\\
		\mu^{(k_L,\ell)}_{i,j}&=& \sigma^{(n,\ell)}_{i,j}\sum_n\left[ y_n^{(\ell)}S_{i,j}^{(n,k_L,L)}(1+\lambda_n^{(\ell)}-\Gamma_{-(k,i,j)}^{(n,k_L,L)})\right],\\
		\Gamma_{-(k,i,j)}^{(n,k_L,L)}&=&\sum_{\substack{k^{\prime}\\ k^{\prime}\neq \ell}}\sum_{\substack{i^{\prime}\\ i^{\prime}\neq i}}\sum_{\substack{j^{\prime}\\ j^{\prime}\neq j}}S_{i^{\prime},j^{\prime}}^{(n,k^{\prime},L)}{\beta}^{(k^{\prime},\ell)}_{i^{\prime},j^{\prime}}.
		\end{eqnarray}	
		\item
		The conditional posterior of $\lambda_n^{(\ell)}$
		\begin{align}\label{Eq:Samplelambda}
		(\lambda_n^{(\ell)})^{-1}\sim {\cal{IG}} (|1-{\yv}_n^{\ell}{\sv}_n^\top\betav^{(\ell,t)}|^{-1}, 1),
		\end{align}
		where ${\cal{IG}}$ denotes the inverse Gaussian distribution.
	\end{enumerate}	
	
\section{MCEM algorithm Details}
	\subsection{E step}
	Recall that we consolidate the ``local'' model parameters (latent data-sample-specific variables) as
	\(\Phimat_n=\big(\{\zv^{(n,l)}\}_{l=1}^L,{\Smat}^{(n,L)},{\gammav}_s^{(n)},{\Emat}^{(n)},\{{\lambda}_n^{(\ell)}\}_{\ell=1}^C\big) ,\) the ``global'' parameters (shared across all data) as \( \Psimat=\big(\{{\Dmat}^{(l)}\}_{l=1}^L,\betav\big),\) and the data as \(\Ymat_n=(\Xmat^{(n)},\ell_n).\). At $t^{\rm th}$ iteration of the MCEM algorithm, 
    the exact $Q$ function can be written as:
    \begin{eqnarray}\label{Eq:exactQ}
    Q(\Psimat|\Psimat^{(t)}) &=&\ln p({\Psimat})+\sum_{n\in\mathcal{I}_t} \mathbb{E}_{(\Phimat_n|\Psimat^{(t)},\Ymat,\yv)}\left\{\ln p({\Ymat}_n,{\Phimat}_n|{\Psimat})\right\}  \nonumber\\   
   &=&-\mathbb{E}_{(\Zmat,\gammav_e,\Smat^{(L)},\gammav_s,\lambdav|\Ymat,\Dmat^{(t)},\betav^{(t)})}\left\{\sum_{n\in \mathcal{I}_t}\left[\frac{{\gamma_e}^{(n)}}{2}\sum_{k_0=1}^{K_0} \|\Emat^{(n,k_0)}\|_2^2+\sum_{\ell=1}^{C}\frac{(1+\lambda_n^{\ell}-y_n^\ell\betav^T_{\ell}\sv_n)^2}{2\lambda_n^{\ell}}\right]\right\} \nonumber\\   
   &&\qquad-\frac{1}{2}\sum_{l=1}^{L}\sum_{k_{l-1}=1}^{K_{l-1}}\sum_{k_l=1}^{K_l}\|{\Dmat}^{(k_{l-1},k_l,l)}\|_2^2   +const,
    \end{eqnarray}
	where $const$ denotes the terms which are not a function of ${\Psimat}$. 
	
	Obtaining a closed form of the exact $Q$ function is analytically intractable. We here approximate the expectations in (\ref{Eq:exactQ}) by samples collected from the posterior distribution of the hidden variables developed in Section \ref{sec:posterior}.\\
    The $Q$ function in (\ref{Eq:exactQ}) can be approximated by:
	\begin{eqnarray}\label{Eq:ApproQ}
	\bar{Q}(\Psimat|\Psimat^{(t)})&=&-\frac{1}{N_s}\sum_{s=1}^{N_s}\left\{\sum_{n\in \mathcal{I}_t}\left[\frac{\bar{\gamma_e}^{(n,s,t)}}{2}\sum_{k_0=1}^{K_0} \|{\bar{\Emat}}^{(n,k_0,s,t)}\|_2^2+\sum_{\ell=1}^{C}\frac{(1+\bar{\lambda_n}^{(\ell,s,t)}-y_n^\ell\betav^T_{\ell}\bar{\sv_n}^{(s,t)})^2}{2\bar{\lambda}_n^{(\ell,s,t)}}\right]\right\}\nonumber\\ &&\quad -\frac{1}{2}\sum_{l=1}^{L}\sum_{k_{l-1}=1}^{K_{l-1}}\sum_{k_l=1}^{K_l}\|{\Dmat}^{(k_{l-1},k_l,l)}\|_2^2   +const,
	\end{eqnarray}\
	where 
	\begin{eqnarray}
	{\bar{\Emat}}^{(n,k_0,s,t)}&=&\Xmat^{(n,k_0)}-\sum_{k_1=1}^{K_1} {\Dmat}^{(k_0,k_1,1)} * \bar {\Smat}^{(n,k_1,1,s,t)},
	\end{eqnarray}
	and for $l=2,\dots,L$
	\begin{eqnarray}
	\bar{\Xmat}^{(n,k_{l-1},l,s,t)}&=&\sum_{k_l=1}^{K_l} {\Dmat}^{(k_{l-1},k_l,l)} * \bar {\Smat}^{(n,k_l,l,s,t)},\\
	\bar{\Smat}^{(n,k_{l-1},l-1,s,t)}&=&f(\bar{\Xmat}^{(n,k_{l-1},l,s,t)}, \bar{\Zmat}^{(n,k_{l-1},l-1,s,t)}),
	\end{eqnarray}
	 where $\bar{\Smat}^{(L,s,t)}$, $\bar{\gammav_e}^{(s,t)}$, $\bar{\lambdav}^{(s,t)}$ and $\bar{\Zmat}^{(s,t)}$ are a sample of the corresponding variables from the full conditional posterior at the $t^{\rm th}$ iteration. $N_s $ is the number of collected samples.
	\subsection{M step} 
	We can maximize $\bar{Q}(\Psimat|\Psimat^{(t)})$ via the following updates:
	\begin{enumerate}
		\item
		For $l=1,\dots,L$, ~$k_{l-1}=1,\dots,K_{L-1}$ and $k_l=1,\dots,K_{L}$, the gradient wrt ${\Dmat}^{(k_{l-1},k_l,l)}$ is:
		\begin{eqnarray}
		    \frac{\partial \bar{Q}}{\partial {\Dmat}^{(k_{l-1},k_l,l,t)}}&=&\sum_{n\in \mathcal{I}_t}\deltav^{(n,k_{l-1},l,t)}\circledast\bar{\Smat}^{(n,k_l,l,t)}+{\Dmat}^{(k_{l-1},k_l,l,t)}, 
		\end{eqnarray}
		where
		\begin{align}\label{Eq:UpdateDelta}
		\begin{array}{rcl}
		\deltav^{(n,k_0,1,t)}&=&\bar{\gamma}_e^{(n,k_0,t)}\left[\Xmat^{(n,k_0)}-\sum_{k_1=1}^{K_1} {\Dmat}^{(k_0,k_1,1)} * \bar {\Smat}^{(n,k_1,1,t)}\right],\\
		\deltav^{(n,k_{l-1},l,t)}&=&f\left(\sum_{k_{l-2}=1}^{K_{l-2}}(\deltav^{(n,k_{l-2},l-1,t)}\circledast D^{(k_{l-2},k_{l-1},l-1,t)}),\bar{Z}^{(n,k_{l-1},l-1,t)}\right).
		\end{array}
		\end{align}
		Following this, the update rule of $\Dmat$ based on RMSprop is:
		\begin{large}
		\begin{align}\label{Eq:UpdateD}
		\begin{array}{rcl}
		\vv^{t+1}&=&\alpha\vv^{t}+(1-\alpha)(\frac{\partial \bar{Q}}{\partial {\Dmat}^{(k_{l-1},k_l,l,t)}})^2,\\
		\\
		{\Dmat}^{(k_{l-1},k_l,l,t+1)}&=&{\Dmat}^{(k_{l-1},k_l,l,t)}+\frac{\epsilon}{\sqrt{\vv_{t+1}}}\frac{\partial \bar{Q}}{\partial {\Dmat}^{(k_{l-1},k_l,l,t)}}.
		\end{array}
		\end{align}
		\end{large}
		\item
		For $\ell=1,\dots,C$, the update rule of $\betav^\ell$ is:
		\begin{align}\label{Eq:UpdateBeta}
		\betav^{(\ell,t+1)}~~=~~\left[(\Omega^{(\ell,t)})^{-1}+\bar{\sv}_{(\ell,t)}^\top (\Lambda^{(\ell,t)})^{-1}\bar{\sv}_{(\ell,t)}\right]^{-1}\bar{\sv}_{(\ell,t)}^\top(\textbf{1}+(\Lambda^{(\ell,t)})^{-1}),
		\end{align}
		where 
		\begin{align}
		(\Lambda^{(\ell,t)})^{-1}&=diag((\bar{\lambda}_n^{(\ell,t)})^{-1}),\\
		(\Omega^{(\ell,t)})^{-1}&=diag(|\betav^{(\ell,t)}|^{-1}).
		\end{align}
		and $\bar{\sv}_{(\ell,t)}$ denotes a matrix with row $n$ equal to ${\yv}_n^{\ell}\bar{\sv}_n^{(t)}$.
	\end{enumerate}
    
   \subsection{Testing}
   During testing, when given a test image ${\Xmat}^{(*)}$, we treat ${\Smat}^{(*,L)}$ as model parameters and use MCEM to find a MAP estimator: 
   \begin{align}
   {\Smat}^{(*,L)}=\underset{{\Smat}^{(*,L)}}{\operatorname{argmax}}\  \ln p({\Smat}^{(*,L)}|{\Xmat}^{(*)},{\Dmat}).
   \end{align}
   Let ${\Smat}^{(*,k_L,L)}={\Wmat}^{(*,k_L,L)}\circ{\Zmat}^{(*,k_L,L)}$, where ${\Wmat}^{(*,k_L,L)}\in \mathbb{R}^{N_{sx}^L\times N_{sy}^L}$.
The marginal posterior distribution can be represented as:
   \begin{eqnarray}
	    p({\Smat}^{(*,L)}|{\Xmat}^{*},{\Dmat})&=& p({\Wmat}^{(*,L)},{\Zmat}^{(*,L)}|{\Ymat}^{(*)},{\Dmat})\\
	   &\propto& \int \sum_{/{\Zmat}^{(L)}} p({\Xmat}^{(*)}|{\Wmat}^{(*,L)},{\Zmat},{\Emat}^{(*)}, {\Dmat})p({\Wmat}^{(*,L)}|\gamma_s^{(*)})p({\Zmat})p(\gamma_s^{(*)})p({\Emat}^{(*)})d{\Emat}^{(*)}d\gamma_s^{(*)}, 
	\end{eqnarray}
   where $/{\Zmat}^{(L)}=\{{\Zmat}^{(l)}\}_{l=1}^{L-1}$. Let $\Psimat_{test}=\{{\Wmat}^{(*,L)},{\Zmat}^{(*,L)}\}$ and $\Phimat_{test}=\{\{{\Zmat}^{(l)}\}_{l=1}^{L-1},{\gammav}_s^{*},{\Emat}^{*}\}$. The $Q$ function for testing can be represented as:
   \begin{align}\label{Eq:testQ}
   & Q_{test}(\Psimat_{test}|\Psimat_{test}^{(t)}) = \mathbb{E}_{(\Phimat_{test}|\Psimat_{test}^{(t)},\,\Ymat^{(*)},\, {\Dmat})}\left\{\ln p({\Xmat}^{(*)},{\Dmat},{\Phimat}_{test}, {\Psimat}_{test})\right\}.
   \end{align}

   The testing also follows EM steps:
     \begin{itemize}
   \item[ E-step:] 
  In the E-step we collect the samples of $\gammav_e$, $\gammav_s$ and $\{{\Zmat}^{(l)}\}_{l=1}^{L-1}$ from conditional posterior distributions, which is similar to the training process. $Q_{test}$ can thus be approximated by:
     \begin{align}\label{Eq:ApproTestQ}
     \bar{Q}_{test}(\Psimat_{test}|\Psimat_{test}^{(t)})&=-\sum_{s=1}^{N_s}\left\{\frac{\bar{\gamma_e}^{(*,s,t)}}{2}\sum_{k_0=1}^{K_0} \|\sum_{k_1=1}^{K_1} {\Dmat}^{(k_{0},k_1,1)} * \bar {\Smat}^{(*,k_1,1,s,t)}\|_2^2+\frac{1}{2}\sum_{k_L=1}^{K_L}\bar{\gamma}_s^{(*,k_L,s)}\|{\Wmat}^{(*,k_L,L)}\|_2^2\right\}
     \end{align}
     where 
     	\begin{align}
     	\bar{\Xmat}^{(*,k_{L-1},L,t)}&=\sum_{k_L=1}^{K_L} {\Dmat}^{(k_{L-1},k_L,L)} *\left( {\Wmat}^{(*,k_L,L)}\circ {\Zmat}^{(*,k_L,L)} \right),
     	\end{align}
  	and for $l=2,\dots,L-1$
  	\begin{align}
  	\bar{\Smat}^{(*,k_{l-1},l-1,s,t)}&=f(\bar{\Xmat}^{(*,k_{l-1},l,t)}, \bar{\Zmat}^{(*,k_{l-1},l-1,s,t)} ),\\
  	\bar{\Xmat}^{(*,k_{l-1},l,s,t)}&=\sum_{k_l=1}^{K_l} {\Dmat}^{(k_{l-1},k_l,l)} * \bar {\Smat}^{(*,k_l,l,s,t)}.
  	\end{align}
  	\item[M-step:]
  	
  	In the M-step, we maximize $\bar{Q}_{test}$ via the following updates:
  		\begin{enumerate}
  		\item 
  		The gradient $w.r.t.$ ${\Wmat}^{(*,K_L,L)}$ is:
  	   \begin{eqnarray}
  	   \frac{\partial \bar{Q}_{test}}{\partial {\Wmat}^{(*,k_L,L,t)}}&=&\left[\sum_{k_{L-1}}^{K_L}\deltav^{(*,k_{L-1},L,t)}\circledast{\Dmat}^{(k_{L-1},k_L,L)}\right]\circ{\Zmat}^{(*,k_L,L)}+\bar{\gamma}_s^{(*,k_L)} {\Wmat}^{(*,k_L,L,t)},
  	   \end{eqnarray}
  	   where $\deltav^{(*,k_{L-1},L,t)}$ is the same as (\ref{Eq:UpdateDelta}). Therefore, the update rule of $\Wmat$ based on RMSprop is:
  	   \begin{large}
  	   	\begin{align}\label{Eq:UpdateW}
  	   	\begin{array}{rcl}
  	   	\uv^{t+1}&=&\alpha\uv^{t}+(1-\alpha)(\frac{\partial \bar{Q}_{test}}{\partial {\Wmat}^{(*,k_L,L,t)}})^2\\
  	   	\\
  	   			{\Wmat}^{(*,K_L,L,t+1)}&=&{\Wmat}^{(*,K_L,L,t)}+\frac{\epsilon}{\sqrt{\uv_{t+1}}}\frac{\partial \bar{Q}_{test}}{\partial {\Wmat}^{(*,k_L,L,t)}}
  	   			\end{array}
  	   	\end{align}
  	  \end{large}
  	   \item
  	   The update rule ${\Zmat}^{(*,k_L,L)}$ is
  	   	\begin{align}\label{Eq:UpdateZ}
  	   	{\Zmat}^{(*,k_L,L)}_{i,j}=\begin{cases}
  	   	1 &\text{ if }\ 	\theta^{(*,k_L,L)}{\eta}^{(*,k_L,L)}_{i,j}>1-\theta^{(*,k_L,L)}\\
  	   	0 & \qquad \qquad\text{ otherwise}
  	   	\end{cases}
  	   	\end{align}
  	   	where ${\eta}^{(*,k_L,L)}_{i,j}$ is the same as (\ref{Eq:eta}). 
  	   	\end{enumerate}
   \end{itemize}

\section{Bottom-Up Pretraining\label{Sec:pre}}
\subsection{Pretraining Model}
The model is pretrained sequentially from the bottom layer to the top layer. We consider here pretraining the relationship between layer $l$ and layer $l+1$, and this process may be repeated up to layer $L$. The basic framework of this pretraining process is closely connected to top-down generative process, with a few small but important modifications. 
Matrix ${\bf X}^{(n, l)}$ represents the pooled and stacked activation weights from layer $l-1$, image $n$ ($K_{l-1}$ ``spectral bands'' in ${\bf X}^{(n,l)}$, due to $K_{l-1}$ dictionary elements at layer $l-1$). We constitute the representation
\begin{align} 	
{\bf X}^{(n, l)} &= {\sum_{k_{l}=1}^{K_{l}} {\bf D}^{(k_{l}, l)}* {\bf S}^{(n,k_{l},l)}  + {\bf E}^{(n, l)}},  
\label{Eq:x_lp1}
\end{align}
with 
\begin{align}
{\bf D}^{(k_{l}, l)}\sim\mathcal{N}(0,\Imat_{N_D^{(l)}})\qquad {\bf E}^{(n,l)} \sim\mathcal{N}(0,(\gammav_e^{(n,l)})^{-1}\Imat_{N_X^{(l)}}) \qquad \gammav_e^{(n,l)}\sim \text{Gamma}(a_e,b_e).
\end{align}
The features ${\bf {S}}^{(n,k_l,l)}$  can be partitioned into contiguous blocks with dimension $p_x^l\times p_y^l$.  In our generative model, ${\bf {S}}^{(n,k_l,l)}$ is generated from ${\bf X}^{(n,k_l,l+1)}$ and $\zv^{(n,k_l,l)}$, where the non-zero element within the $(i,j)-$th pooling block of ${\bf {S}}^{(n,k_l,l)}$ is set equal to $X_{i,j}^{(n,k_l,l+1)}$, and its location within the pooling block is determined by $\zv^{(n,k_l,l)}_{i,j}$, a $p_x^l\times p_y^l$ binary vector (Sec. 2.2 in the main paper). Now the matrix ${\bf X}^{(n,k_l,l+1)}$ is constituted by ``stacking'' the spatially-aligned and pooled versions of ${\bf {S}}^{(n,k_l,l)}_{k_l=1,K_{l}}$. Thus, we need to place a prior on the $(i,j)-$th pooling block of ${\bf {S}}^{(n,k_l,l)}$:  
\begin{align}
{\bf {S}}_{i,j,m}^{(n,k_l,l)}&=z^{(n,k_l,l)}_{i,j,m}W^{(n,k_l,l)}_{i,j,m}, \quad m=1,\dots,p_x^lp_y^l&&\\
\zv^{(n,k_l,l)}_{i,j}&\sim \thetav^{(n,k_l,l)}_0[\zv^{(n,k_l,l)}_{i,j}={\bf 0}]+\sum_{m=1}^{p_x^lp_y^l}\thetav^{(n,k_l,l)}_m [\zv^{(n,k_l,l)}_{i,j}={\ev_m}],&\thetav^{(n,l,k_l)} &\sim \textstyle{\mbox{Dir}} (1/p_x^lp_y^l,\dots,1/p_x^lp_y^l), \\
W^{(n,k_l,l)}_{i,j,m}&\sim \mathcal{N}(0,\gamma_{wl}^{-1}), &\gamma_{wl}&\sim\text{Gamma}(a_w,b_w).
\end{align}
If  all the elements of $\zv^{(n,l)}_{i,j,k_l}$ are zero, the corresponding pooling block in ${\bf S}_{i,j}^{(n,k_l,l)}$ will be all zero and $X_{i,j}^{(n,k_l,l+1)}$ will be zero.

Therefore, the model can be formed as:	
\begin{align}
{\Xmat}^{(n, l)} = \sum_{k_{l}=1}^{K_{l}} {\Dmat}^{(k_{l}, l)} *\underbrace{\left({\Zmat}^{(n,k_{l},l)}  \odot {\Wmat}^{(n,k_{l}, l)}\right)}_{={ \Smat}^{(n,k_{l},l)}}  + {\Emat}^{(n, l)}, \label{Eq:x_lp1}
\end{align}
where the vector version of the $(i,j)$-th block of ${\bf {Z}}^{(n,k_l,l)}$ is equal to  $\zv^{(n,k_l,l)}_{i,j}$ and $\odot$ is the Hadamard (element-wise) product operator. The hyperparameters are set as $a_e=b_e=a_w=b_w=10^{-6}$.

We summarize distinctions between pretraining, and the top-down generative model. 
\begin{itemize}
	\item 
	A pair of consecutive layers is considered at a time during pretraining.
	\item 
	During the pretraining process, the residual term $ {\bf E}^{(n, l)}$ is used to fit each layer.
	\item
	In the top-down generative process, the residual is only employed at the bottom layer to fit the data.
	\item 
	During pretraining, the pooled activation weights  ${\bf X}^{(n,l+1)}$ are sparse, encouraging a parsimonious convolutional dictionary representation. 
	\item 
	The model parameters learned from pretraining are used to initialize the model when executing top-down model refinement, using the full generative model.
\end{itemize} 
\subsection{Conditional Posterior Distribution for Pretraining}
\begin{itemize}
	\item
	$\displaystyle D^{(k_{l-1},k_l,l)}_{i,j}|- \sim {\mathcal N}({\Phimat}^{(k_{l-1},k_l,l)}_{i,j},\Sigmamat_{i,j}^{(k_{l-1},k_l,l)})$
	\begin{align}
	\Sigmamat^{(k_{l-1},k_l,l)} &= \textbf{1} \oslash \left(\sum_{n=1}^{N} \gamma_e^{(n,l)} \|{\Zmat}^{(n,k_{l},l)} \odot {\Wmat}^{(n,k_{l},l)}\|_2^2 + \textbf{1}\right)\\
	{\Phimat}^{(k_{l-1},k_l,l)} &= \Sigmamat^{(k_{l-1},k_l,l)} \odot \Bigg\{\sum_{n=1}^{N} \gamma_e^{(n,l)} \Big[{\Xmat}^{-(n,k_{l-1},l)} \circledast ({\Zmat}^{(n,k_{l},l)} \odot {\Wmat}^{(n,k_{l},l)})\nonumber\\
	&\qquad+\|{\Zmat}^{(n,k_{l},l)} \odot {\Wmat}^{(n,k_{l},l)}\|_2^2 {\Dmat}^{(k_{l-1},k_l,l)}\Big]\Bigg\}
	\end{align}
	\item
	$\displaystyle W_{i,j}^{(n,k_{l},l)}|- \sim {\mathcal N}(\Ximat_{i,j}^{(n,k_{l},l)}, \Lambdamat_{i,j}^{(n,k_{l},l)})\\$
	\begin{align}
	\Lambdamat^{(n,k_{l},l)} &= \textbf{1} \oslash \left(\sum_{k_{l-1}=1}^{K_{l-1}} \gamma_e^{(n,l)} \|{\Dmat}^{(k_{l-1},k_l,l)} \|_2^2 {\Zmat}^{(n,k_{l},l)} + \gamma_w^{(n,k_{l},l)}\textbf{1}\right)\\
	\Ximat^{(n,k_{l},l)} &= \Lambdamat^{(n,k_{l},l)} \odot {\Zmat}^{(n,k_{l},l)} \odot \Bigg\{\sum_{k_{l-1}=1}^{K_{l-1}} \gamma_e^{(n,l)} \Big[{\Xmat}^{-(n,k_{l-1},l)}  \circledast {\Dmat}^{(k_{l-1},k_l,l)}\nonumber\\
	&\qquad+\|{\Dmat}^{(k_{l-1},k_l,l)} \|_2^2 {\Wmat}^{(n,k_{l},l)} \odot{\Zmat}^{(n,k_{l},l)}\Big]\Bigg\}
	\end{align}
	\item
	$\displaystyle\gamma_w^{(n,k_{l},l)}|- \sim \text{Gamma}\left(a_w+\frac{N_{sx}^l\times N_{sy}^l}{2}, b_w+\frac{\|{\Wmat}^{(k_{l-1},k_l,l)} \|_2^2}{2}\right)$
	\item
	$\zv^{(n,k_l,l)}_{i,j}$:\\

	Let $m\in\{1,...,p_x^lp_y^l\}$; from
	\begin{align}
	{\Ymat}^{(n,k_{l},l)} = \sum_{k_{l-1}=1}^{K_{l-1}}& \gamma_e^{(n,l)} \big[ \|{\Dmat}^{(k_{l-1},k_{l},l)}\|_2^2 \odot \left({\Wmat}^{(n,k_{l},l)}\right)^2 - 2 \left( {\Xmat}_{-k_l}^{(n,k_{l-1},l)} \circledast {\Dmat}^{k_{l-1},k_l,l} \right) \odot {\Wmat}^{(n,k_{l},l)} \big]
	\end{align}
	and
	\begin{align}
	\hat{\theta}_{i,j,m}^{(n,k_{l},l)} &= \frac{\theta_m^{(n,k_{l},l)} \exp\left\{-\frac{1}{2}Y_{i,j,m}^{(n,k_{l},l)}\right\}}{\theta_0^{(n,k_{l},l)}+\sum_{\hat{m}=1}^{p_x^lp_y^l}\theta_{\hat{m}}^{(n,k_{l},l)} \exp\left\{-\frac{1}{2}Y_{i,j,{\hat{m}}}^{(n,k_{l},l)}\right\}},\\
	\hat{\theta}_{i,j,0}^{(n,k_{l},l)} &= \frac{\theta_0^{(n,k_{l},l)} }{\theta_0^{(n,k_{l},l)}+\sum_{\hat{m}=1}^{p_x^lp_y^l}\theta_{\hat{m}}^{(n,k_{l},l)} \exp\left\{-\frac{1}{2}Y_{i,j,{\hat{m}}}^{(n,k_{l},l)}\right\}},\\
	\end{align}
	we have
	\begin{align}
	\zv^{(n,k_l,l)}_{i,j}|&\sim \hat{\thetav}^{(n,k_l,l)}_0[\zv^{(n,k_l,l)}_{i,j}={\bf 0}]+\sum_{m=1}^{p_x^lp_y^l}\hat{\thetav}^{(n,k_l,l)}_m[\zv^{(n,k_l,l)}_{i,j}={\ev_m}].
	\end{align}
	\item
	$\displaystyle\thetav^{(n,k_{l},l)} |- \sim \text{Dir} ({\alpha}^{(n,k_{l},l)})$
	\begin{align}
	\alpha_m^{(n,k_{l},l)} &= \frac{1}{p_x^lp_y^l+1} + \sum_{i} \sum_{j} Z_{i,j,m}^{(n,k_{l},l)} \qquad \text{for }~ m=1,...,p_x^lp_y^l,\\
	\alpha_{0}^{(n,k_{l},l)} &= \frac{1}{p_x^lp_y^l+1} + \sum_{i} \sum_{j} \left(1-\sum_m Z_{i,j,m}^{(n,k_{l},l)}\right)
	\end{align}
	\item
	$\displaystyle\gamma_e^{(n,l)}|- \sim \text{Gamma}\left(a_e+\frac{N_x^l\times N_y^l\times K_{l-1}}{2}, b_e+\sum_{k_{l-1}=1}^{K_{l-1}}\frac{\|{\Xmat}^{-(n,k_{l-1},l)} \|_2^2}{2}\right)$
\end{itemize}
\end{document}